\documentclass[journal]{IEEEtran}

\usepackage{graphicx}
\usepackage{multirow}
\usepackage{booktabs}
\usepackage{tikz}
\usepackage{comment} 
\usepackage{amsmath,amssymb} 
\usepackage{color}
\usepackage{caption}

\usepackage{hyperref}

\begin{document}

\title{Image-to-Image Translation: Methods and Applications}

\author{Yingxue Pang, Jianxin Lin, Tao Qin,~\IEEEmembership{Senior Member,~IEEE,} and
Zhibo Chen\IEEEauthorrefmark{1},~\IEEEmembership{Senior Member,~IEEE,}
\thanks{Yingxue Pang, Jianxin Lin, and Zhibo Chen are with the Department of Electronic Engineer and Information Science, University of Science and Technology of China, Hefei, Anhui, 230026, China. (e-mail:~pangyx@mail.ustc.edu.cn;~linjx@mail.ustc.edu.cn;~chenzhibo@ustc.edu.cn.)}
\thanks{Tao Qin are with Microsoft Research Asia. (e-mail:taoqin@microsoft.com)}
\thanks{\IEEEauthorrefmark{1} Corresponding author.}
}

\maketitle

\begin{abstract}
Image-to-image translation (I2I) aims to transfer images from a source domain to a target domain while preserving the content representations. I2I has drawn increasing attention and made tremendous progress in recent years because of its wide range of applications in many computer vision and image processing problems, such as image synthesis, segmentation, style transfer, restoration, and pose estimation. In this paper, we provide an overview of the I2I works developed in recent years. We will analyze the key techniques of the existing I2I works and clarify the main progress the community has made. Additionally, we will elaborate on the effect of I2I on the research and industry community and point out remaining challenges in related fields.
\end{abstract}


\IEEEpeerreviewmaketitle

\section{Introduction}
\label{sec:intro}
\begin{figure}[!t]
\centering
\includegraphics[width=0.45\textwidth]{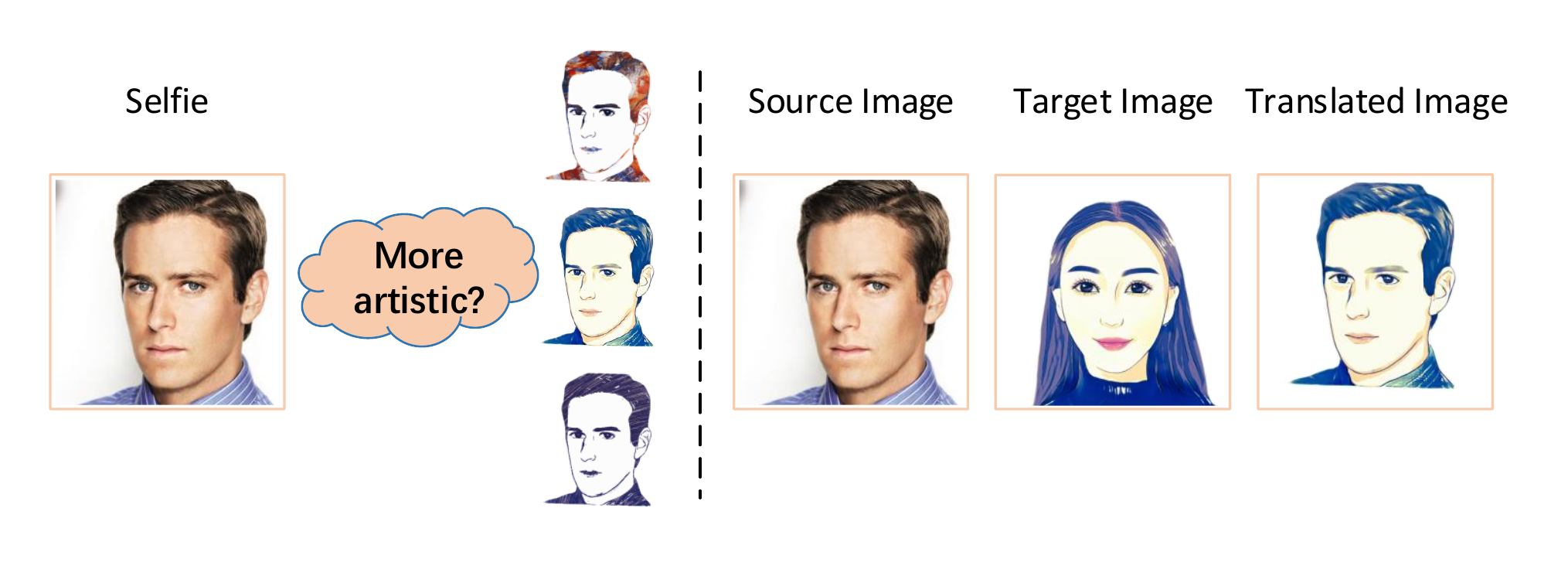}
\caption{An example of image-to-image translation (I2I) for illustration. (Left): How to make your selfie more artistic as drawings from cartoonists?  This type of research work can be broadly deemed the I2I problem. (Right): You can take a selfie as a source image and a cartoon as a target reference to ``translate" it into desired artistic style image.}
\label{fig:selfie}
\end{figure}

\IEEEPARstart{I}{magine} 
if you take a selfie and want to make it more artistic as a drawing from a cartoonist, 
how can you automatically achieve that with a computer? This type of research work can be broadly deemed the image-to-image translation (I2I) (\cite{isola2017image,zhu2017unpaired}) problem. In general, the goal of I2I is to convert an input image $x_{A}$ from a source domain $A$ to a target domain $B$ with the intrinsic source content preserved and the extrinsic target style transferred. For example, one can take selfie images as the source domain and ``translate" them to desired artistic style images given some cartoons as target domain references, as shown in Fig. \ref{fig:selfie}. 
To this end, we need to train a mapping $G_{A\mapsto{B}}$ that generates image $x_{AB}\in{B}$ indistinguishable from target domain image $x_{B}\in{B}$ given the input source image $x_{A}\in{A}$. Mathematically, we can model this translation process as
\begin{equation}
\label{eq:pro_setting}
x_{AB}\in{B}: x_{AB} = G_{A\mapsto{B}}(x_A).
\end{equation}

From the above basic definition of I2I, we see that converting an image from one source domain to another target domain can cover many problems in image processing, computer graphics, computer vision and so on. Specifically, I2I has been broadly applied in semantic image synthesis \cite{regmi2018cross,Park_2019_CVPR,Zhu_2020_CVPR,lee2020maskgan,Tang_2020_CVPR}, image segmentation \cite{yang2018mri,GUO2020127,LI2020107343}, style transfer \cite{zhu2017unpaired,kim2017learning,yi2017dualgan,mejjati2018unsupervised,tomei2019art2real}, image inpainting \cite{Pathak_2016_CVPR,zhu2017toward,Song_2018_ECCV,Liu_2019_ICCV,Zhao_2020_CVPR}, 3D pose estimation \cite{Tung_2017_ICCV,Li_2020_CVPR}, image/video colorization \cite{zhang2017real,suarez2017infrared,he2018deep,zhang2019deep,xu2020stylization,Lee_2020_CVPR}, image super-resolution \cite{Yuan_2018_CVPR_Workshops,zhang2019multiple}, domain adaptation \cite{Murez_2018_CVPR,cao2018dida,liu2018unified}, cartoon generation \cite{shi2019warpgan,pkesko2019comixify,zheng2019unpaired,chen2018cartoongan,kim2019u,wang2020learning} and image registration \cite{Arar_2020_CVPR}. We will analyze and discuss these related applications in detail in Section \ref{sec:application}.

\begin{figure*}[!t]
\centering
\includegraphics[width=\textwidth]{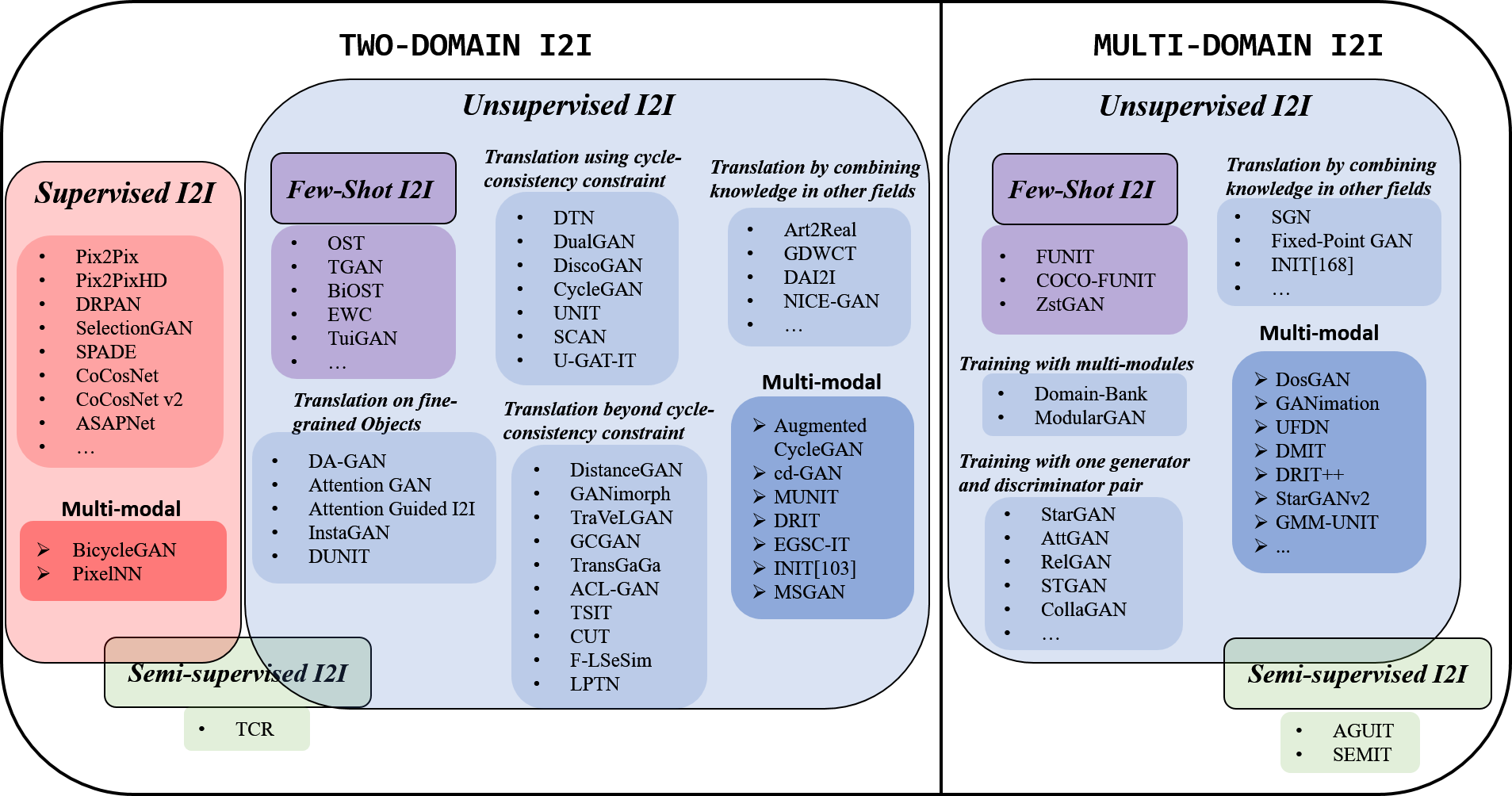}
\caption{An overview of image-to-image translation methods. This figure shows the relationship between different methods and where they intersect with each other. }
\label{fig:whole1}
\end{figure*}

In this paper, we aim to provide a comprehensive review of the recent progress in image-to-image translation research works. To the best of our knowledge, this is the first overview paper to cover the analysis, methodology, and related applications of I2I. In detail, we present our survey with the following organization:
\begin{itemize}
\item First, we briefly introduce the two most representative and commonly adopted generative models, as well as some well-known evaluation metrics, applied for image-to-image translation, and then we analyze how these generative models learn to represent and acquire the desired translation results.
\item Second, we categorize the I2I problem into two main sets of tasks, i.e., two-domain I2I tasks and multi-domain I2I tasks, in which numerous I2I works have appeared for each set of I2I tasks and brought far-reaching influence on other research fields, as shown in Fig. \ref{fig:whole1}.
\item Last but not least, we provide a thorough taxonomy of the I2I applications following the same categorizations of I2I methods, as illustrated in Table. \ref{table:UI2I}.
\end{itemize}
In general, our paper is organized as follows. Section \ref{sec:intro} provides the problem setting of the image-to-image translation task. Section \ref{sec:backbone} introduces the generative models used for I2I methods. Section \ref{sec:two-domain} discusses the works on the two-domain I2I task. Section \ref{sec:multi-domain} focuses on works related to the multi-domain I2I task. Then, Section \ref{sec:application} reviews the various and fruitful applications of I2I tasks. Summary and outlook are given in Section \ref{sec:conclusion}.

\section{The Backbone of I2I}
\label{sec:backbone}
Because an I2I task aims to learn the mappings between different image domains, how to represent these mappings to generate the desirable results is explicitly related to the generative models. The generative model \cite{lecun2006tutorial,xu2015overview,oussidi2018deep} assumes that data is created by a particular distribution that is defined by two parameters (i.e., a Gaussian distribution) or non-parametric variants (each instance has its own contribution to the distribution), and it approximates that underlying distribution with particular algorithms. This approach enables the generative model to generate data rather than only discriminate between data (classification). For instance, the deep generative models have shown substantial performance improvements in making predictions \cite{chu2018deep}, estimating missing data \cite{yeh2017semantic}, compressing datasets \cite{tschannen2018deep} and generating invisible data. In an I2I task, a generative model can model the distribution of the target domain by producing convincing ``fake" data, namely, the translated images, that appear to be drawn from the distribution of the target domain. 

However, considering the length of this article and the difference in research foci, we inevitably omit those generative models that are vaguely connected with the theme of I2I, such as deep Boltzmann machines (DBMs) \cite{salakhutdinov2009deep,salakhutdinov2007restricted,hinton2006fast}, deep autoregressive models (DARs) \cite{van2016conditional,van2016pixel,chen2018pixelsnail} and normalizing flow models (NFMs) \cite{dinh2014nice,rezende2016variational,Abdelhamed_2019_ICCV}. Therefore, we will briefly introduce two of the most commonly used and efficient deep generative models in I2I tasks, variational autoencoders (VAEs) \cite{dayan1995helmholtz,kingma2014auto,rezende2014stochastic,larochelle2011neural,germain2015made,dinh2014nice,nalisnick2016approximate,rezende2015variational,tomczak2018vae,tolstikhin2018wasserstein} and generative adversarial networks (GANs) \cite{goodfellow2014generative,arjovsky2017wasserstein,mao2017least,radford2015unsupervised,mirza2014conditional,gulrajani2017improved,zhao2016energy,berthelot2017began,jolicoeur2018relativistic,miyato2018spectral}, as well as the intuition behind them. Both models basically aim to construct a replica $x=g(z)$ for generating the desired samples $x$ from the latent variable $z$, but their specific approaches are different. A VAE models data distribution by maximizing the lower bound of the data log-likelihood, whereas a GAN tries to find the Nash equilibrium between a generator and discriminator.

On the other hand, after obtaining the translated results from the generative model, we need subjective and objective metrics for evaluating the quality of translated images. Therefore, we will also briefly present common evaluation metrics in the I2I problem.

\subsection{Variational AutoEncoder}
Inspired by the Helmholtz machine \cite{dayan1995helmholtz}, the variational autoencoder (VAE) \cite{kingma2014auto,rezende2014stochastic} was initially proposed for a variational inference problem in deep latent Gaussian models. 

\begin{figure}[!t]
\centering
\includegraphics[width=0.45\textwidth]{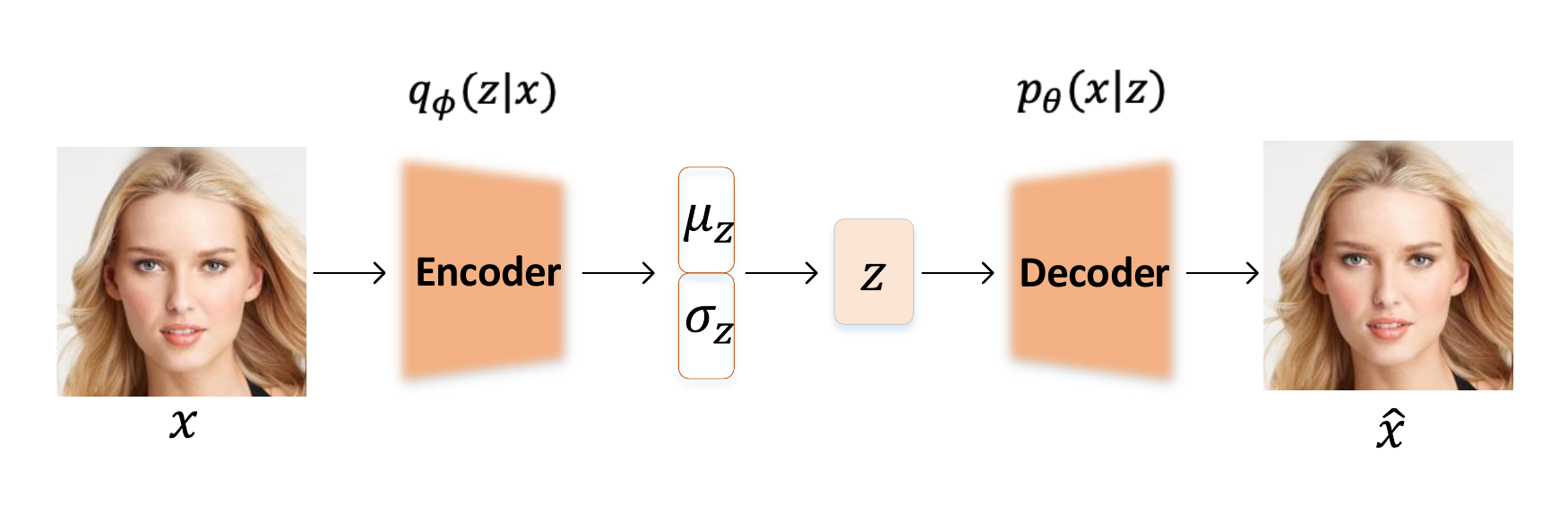}
\caption{The structure of a VAE}
\label{fig:vae}
\end{figure}

As shown in Fig \ref{fig:vae}, a VAE \cite{kingma2014auto,rezende2014stochastic} adopts a recognition model (encoder) $q_{\phi}(z|x)$ to approximate the posterior distribution $p(z|x)$ and a generative model (decoder) $p_{\theta}(x|z)$ to map the latent variable $z$ to the data $x$. Specifically, a VAE trains its generative model to learn a distribution $p(x)$ to be near the given data $x$ by maximizing the log-likelihood function $logp_{\theta}(x)$:
\begin{equation}
\begin{aligned}
&logp_{\theta}(x)=\sum_{i=1}^{N}logp_{\theta}(x_{i}),\\
&logp_{\theta}(x_{i})=log\int{p_{\theta}(x_{i}|z)p(z)dz}.\qquad
\end{aligned}
\label{eq:vae1}
\end{equation}
Stochastic gradient ascent (SGA) combined with the naive Monte Carlo gradient estimator (MCGE) can be used to find the optimal solution in Eqn.\eqref{eq:vae1}. However, it often fails because of the highly skewed samples $p_{\theta}(x|z)$ that exhibit a very high variance. A VAE therefore introduces the recognition model $q_{\phi}(z|x)$ as a multivariate Gaussian distribution with a diagonal covariance structure:
\begin{equation}
q_{\phi}(z|x)= \mathcal{N}(z|\mu_{z}(x,\phi),\sigma_{z}^{2}(x,\phi)I).
\end{equation}
Eqn.\eqref{eq:vae1} can be rewritten as:
\begin{equation}
logp_{\theta}(x_{i})=\mathcal{L}(x_{i},\theta,\phi)+D_{KL}[q_{\phi}(z|x_{i})||p_{\theta}(z|x_{i})].
\end{equation}
where $D_{KL}$ denotes the KL divergence that is non-negative, and $\theta$ and $\phi$ are neural network parameters. Naturally, we can obtain a variational lower bound on the log-likelihood:
\begin{equation}
logp_{\theta}(x_{i})\ge{\mathcal{L}(x_{i},\theta,\phi)}.
\end{equation}
Hence, a VAE differentiates and optimizes the lower bound $\mathcal{L}(x_{i},\theta,\phi)$ instead of $logp_{\theta}(x_{i})$. Here is the final objective function of a VAE:
\begin{equation}
\begin{aligned}
\mathcal{L}(x_{i},\theta,\phi)=E_{z\sim{q_{\phi}(z|x_{i})}}[logp_{\theta}(x_{i}|z)]-\\
D_{KL}[q_{\phi}(z|x_{i})||p_{\theta}(z|x_{i})].
\end{aligned}
\end{equation}

As stated in \cite{ghosh2019variational}, VAEs provide more stable training than generative adversarial networks (GANs) \cite{goodfellow2014generative} and more efficient sampling mechanisms than autoregressive models \cite{larochelle2011neural,germain2015made}. However, several practical and theoretical challenges of VAEs remain unsolved. 
The main drawback of variational methods is their tendency to strike an unsatisfactory trade-off between the sample quality and the reconstruction quality because of the weak approximate posterior distribution or overly simplistic posterior distribution. The studies in \cite{dinh2014nice,nalisnick2016approximate,rezende2015variational} enrich the variational posterior to alleviate the blurriness of generated samples. Tomczak et al. \cite{tomczak2018vae} proposed a new prior, VampPrior, to learn more powerful hidden representations. In addition, \cite{tolstikhin2018wasserstein} claimed that the inherent over-regularization induced by the KL divergence term in the VAE objective 
often leads to a gap between $\mathcal{L}(x_{i},\theta,\phi)$ and the true likelihood.

Generally, with the development of VAEs, this type of generative model constitutes one well-established approach for I2I tasks \cite{zhu2017toward,liu2017unsupervised,lee2018diverse,ma2018exemplar,wu2019transgaga}. Next, we will introduce another important generative model, generative adversarial networks, which have been widely used in multiple I2I models \cite{isola2017image,kazemi2018unsupervised,zhu2017unpaired,lin2019zstgan,lin2020tuigan}.

\subsection{Generative Adversarial Networks}
The main idea of generative adversarial networks (GANs) \cite{goodfellow2014generative,arjovsky2017wasserstein,mao2017least} is to establish a zero-sum game between two players, namely, a generator and discriminator, in which each player is represented by a differentiable function controlled by a set of parameters. Generator $G$ tries to generate fake but plausible images, while discriminator $D$ is trained to distinguish the difference between real and fake images. The solution of this game is to find a Nash equilibrium between the two players. In the following subsection, we will discuss the unconditional GANs, the conditional GANs and the way to train GANs. 

\subsubsection{Unconditional GANs}
\begin{figure}[!t]
\centering
\includegraphics[width=0.45\textwidth]{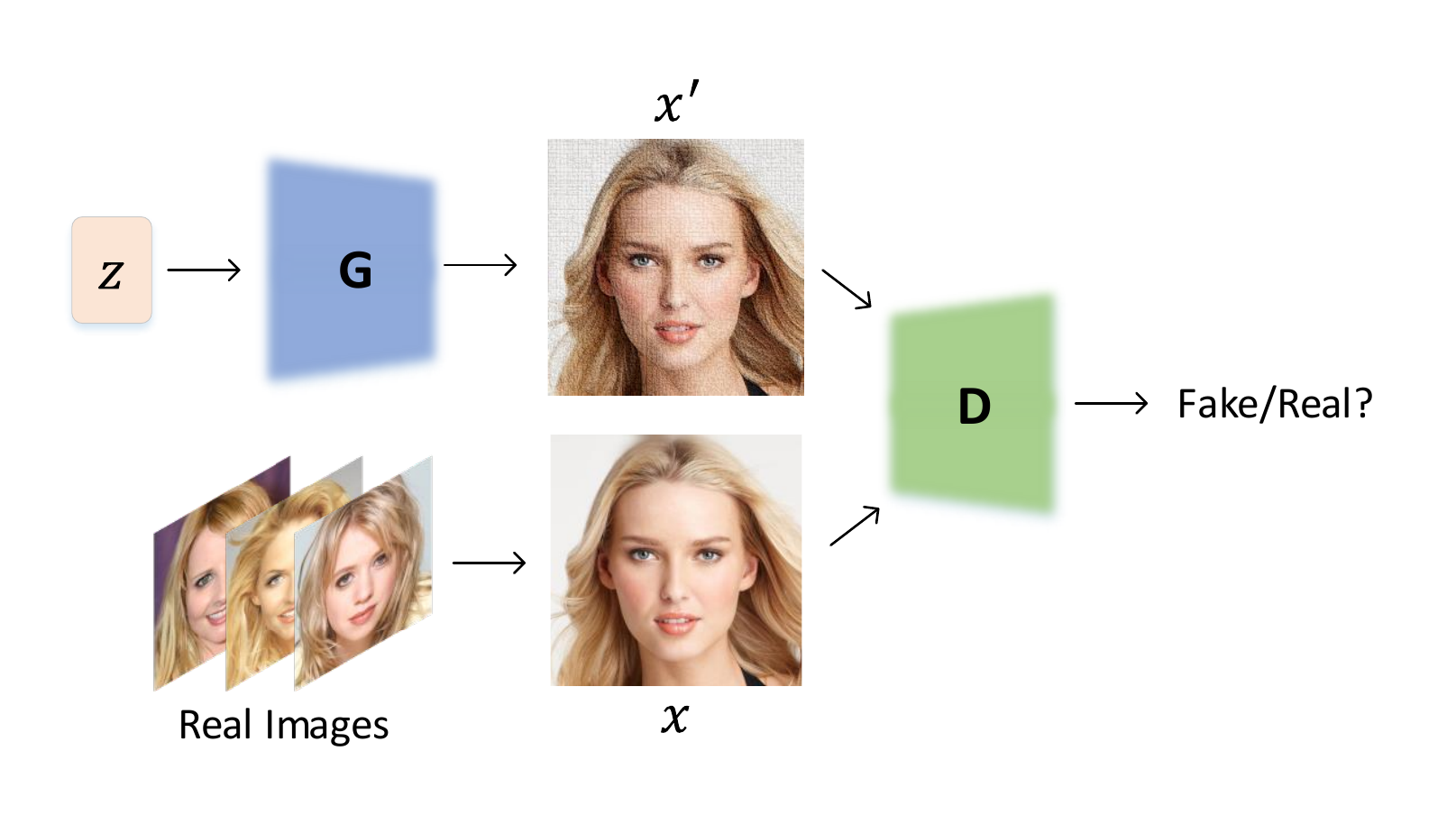}
\caption{The structure of unconditional GANs, where $z, G$ and $D$ denote the random noise, generator, and discriminator, respectively.}
\label{fig:uncondi_gan}
\end{figure}
The original GAN proposed by \cite{goodfellow2014generative} can be considered an unconditional GAN. It adopts the multilayer perceptron (MLP) \cite{pal1992multilayer} to construct a structured probabilistic model taking latent noise variables $z$ and observed real data $x$ as inputs. Because the convolutional neural network (CNN) \cite{lecun1989backpropagation} has been demonstrated to be more effective than the MLP in representing image features, the studies in \cite{radford2015unsupervised} proposed the deep convolutional generative adversarial networks (DGANs) to learn a better representation of images and improve the original GAN performance.

As illustrated in Fig \ref{fig:uncondi_gan}, the generator $G$ inputs a random noise $z$ sampled from the model's prior distribution $p(z)$ to generate a fake image $G(z)$ to fit the distribution of real data as much as possible. Then, the discriminator $D$ randomly takes the real sample $x$ from the dataset and the fake sample $G(z)$ as input to output a probability between 0 and 1, indicating whether the input is a real or fake image. In other words, $D$ wants to discriminate the generated fake sample $G(z)$ while $G$ intends to create samples to confuse $D$.
Consequently, the objective optimization problem is as shown below:
\begin{equation}
\begin{aligned}
\mathop{min}\limits_{G}\mathop{max}\limits_{D}\mathcal{L}(D,G)=E_{x\sim{p_{data}(x)}}[logD(x)]+\\
E_{z\sim{p_{z}(z)}}[log(1-D(G(z)))].
\end{aligned}
\end{equation}
where $x$ denotes the real data, $z$ denotes the random noise vector, and $G(z)$ are the fake samples generated by the generator $G$. $D(x)$ indicates the probability that $D$'s input is real, and $D(G(z))$ denotes the probability that $D$ discriminates between the input generated by $G$.

\begin{figure}[!t]
\centering
\includegraphics[width=0.45\textwidth]{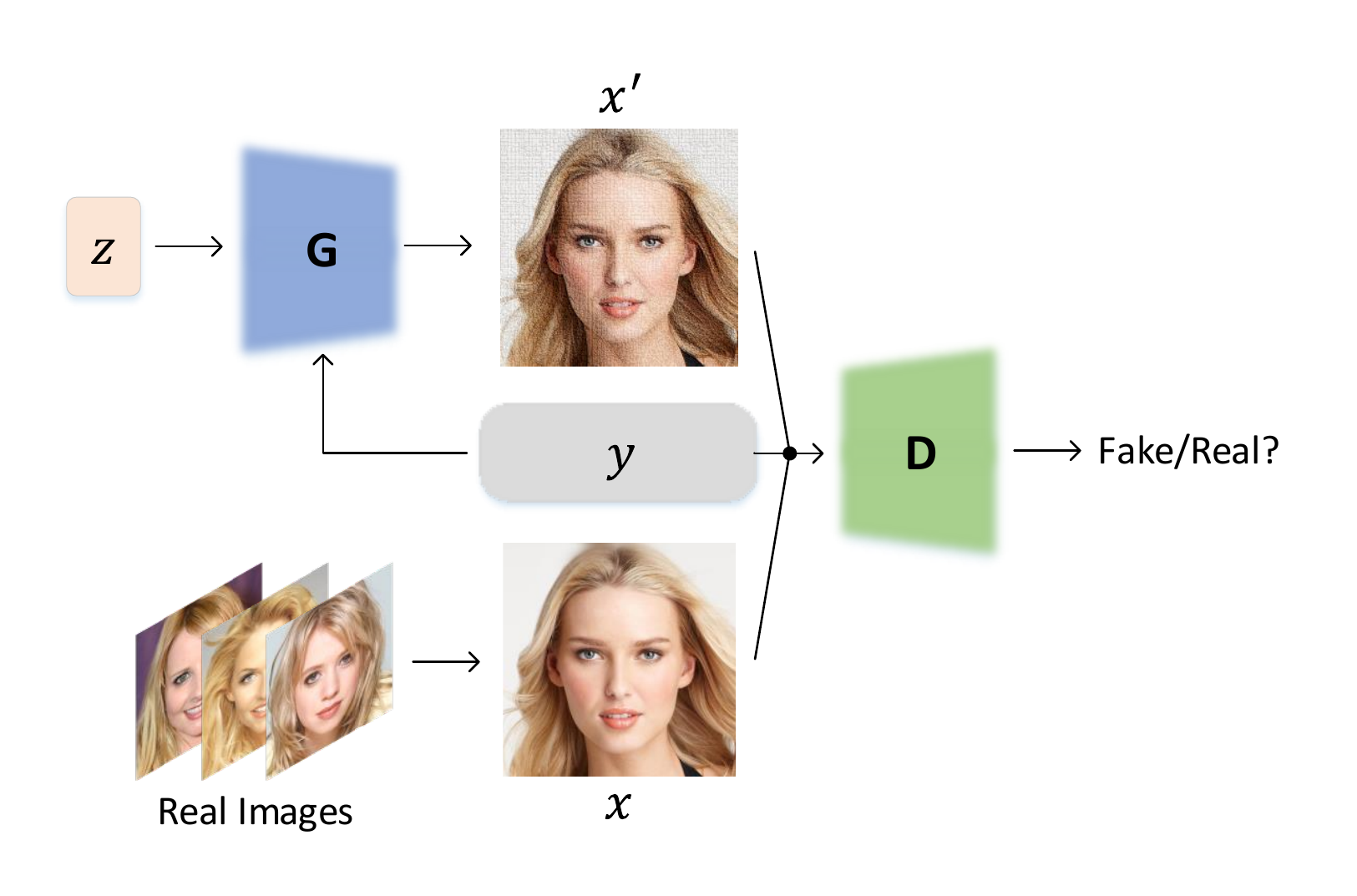}
\caption{The structure of conditional GANs, where $z, G$ and $D$ denote the random noise, generator, and discriminator, respectively. Conditional GANs usually add additional information $y$ (such as data labels, text or attributes of images) to the generator and discriminator to generate desirable results.}
\label{fig:condi_gan}
\end{figure}
\subsubsection{Conditional GANs} 
In the unconditional GAN, there is no control of what we want to generate because the only input is the random noise vector $z$. Therefore, \cite{mirza2014conditional} proposed adding additional information $y$ concatenated with $z$ to generate image $G(z|y)$ shown in Fig \ref{fig:condi_gan}. The conditional input $y$ can be any information, such as data labels, text and attributes of images. In this way, we can use the additional information to adjust the generated results in a desirable direction. The objective function is described as:
\begin{equation}
\begin{aligned}
\mathop{min}\limits_{G}\mathop{max}\limits_{D}\mathcal{L}(D,G)=E_{x\sim{p_{data}(x)}}[logD(x|y)]+\\
E_{z\sim{p_{z}(z)}}[log(1-D(G(z|y)))].
\end{aligned}
\end{equation}
Note that the real data is also under the control of the same conditional variable $y$, i.e., $D(x|y)$.

\subsubsection{The Way to Train GANs}
In the training process, GAN updates the parameters of $G$ along with $D$ using gradient-based optimization methods, such as stochastic gradient descent (SGD), Adam \cite{kingma2015adam} and RMSProp \cite{Tieleman2012}. The entire optimization goal is achieved when $D$ cannot distinguish between the generated sample $x'=G(z)$ and real sample $x$, i.e., when the Nash equilibrium is found in this status. In practice, the training of GANs is often trapped in mode collapse, and it is difficult to achieve convergence. 

To address the stability problem, many recent studies focus on finding new cost functions with smoother non-vanishing or non-exploding gradients everywhere. WGAN\cite{arjovsky2017wasserstein} proposes a new cost function using the Wasserstein distance to address the mode collapse problem appearing in naive GAN \cite{goodfellow2014generative}, and WGAN-GP\cite{gulrajani2017improved} uses a gradient penalty instead of the weight clipping to enforce the Lipschitz constraint in WGAN. LSGAN\cite{mao2017least} finds that optimizing the least squares cost function is identical to optimizing a Pearson $\chi^2$ divergence. EBGAN\cite{zhao2016energy} replaces the discriminator with an autoencoder and uses the reconstruction cost (MSE) to criticize the real and generated images. BEGAN\cite{berthelot2017began} builds with the same EBGAN autoencoder concept for the discriminator but with different cost functions. RSGAN\cite{jolicoeur2018relativistic} measures the probability that the real data is more realistic than the generated data, making the cost function relativistic. SNGAN\cite{miyato2018spectral} proposes a weight normalization technique called spectral normalization to stabilize the training of the discriminator.

\subsection{Evaluation Metrics}
To reflect the visual quality of the translation performance more comprehensively, we also introduce some common evaluation metrics used in I2I, including subjective and objective metrics.

\subsubsection{Subjective image quality assessment}
\begin{itemize}
\item \textbf{AMT perceptual studies}: 
This test is a ``real or fake" two-alternative forced choice experiment on the Amazon Mechanical Turk (AMT) used in many I2I algorithms \cite{zhang2016colorful,isola2017image,zhu2017unpaired,Wang_2018_ECCV}. Turkers are presented a series of pairs of images, one real and one fake (generated by the I2I models). Participants are asked to choose the photo they think is real and then obtain the feedback to compute the scores.
\end{itemize}

\subsubsection{Objective image quality assessment}
\begin{itemize}
\item \textbf{Peak signal-to-noise ratio (PSNR)}: PSNR is one of the most widely used full-reference quality metrics. 
It reflects the intensity differences between the translated image and its ground truth. A higher PSNR score means that the intensity of two images is closer.
\item \textbf{Structural similarity index (SSIM) \cite{wang2004image}}: 
I2I uses SSIM to compute the perceptual distance between the translated image and its ground truth. The higher the SSIM is, the greater the similarity of the luminance, contrast and structure of two images will be.
\item \textbf{Inception score (IS)} \cite{salimans2016improved}: IS encodes the diversity across all translated outputs. It exploits a pretrained inception classification model to predict the domain label of the translated images. A higher score indicates a better translated performance.
\item \textbf{Mask-SSIM and Mask-IS} \cite{ma2017pose}: These two metrics are the masked versions of SSIM and IS to reduce the background influence by masking it out. They are designed for evaluating the performance of person image generation task.
\item \textbf{Conditional inception score (CIS)} \cite{huang2018multimodal}: CIS is modified from IS to better evaluate the multimodal I2I works. It encodes the diversity of the translated output conditioned on a single input image. A higher score indicates a better translated performance.
\item \textbf{Perceptual distance (PD)} \cite{johnson2016perceptual}: PD computes the perceptual distance between the translated image and corresponding source image. A lower PD score indicates that the contents of two images are more similar.
\item \textbf{Fr\'{e}chet inception distance (FID)} \cite{heusel2017gans}: The FID measures the distance between the distributions of synthesized images and real images. A lower FID score means a better performance.
\item \textbf{Kernel inception distance (KID)} \cite{binkowski2018demystifying}: The KID computes the squared maximum mean discrepancy between the feature representations of real and generated images in which feature representations are extracted from the Inception network \cite{szegedy2016rethinking}. A lower KID indicates more shared visual similarities between real and generated images.
\item \textbf{Single image Fr\'{e}chet inception distance (SIFID)} \cite{shaham2019singan}: The SIFID captures the difference between the internal distributions of two images, which is implemented by computing the Fr\'{e}chet inception distance (FID) between the deep features of two images. The SIFID is computed using the translated image and corresponding target image. A lower SIFID score indicates that the styles of two images are more similar. 
\item \textbf{LPIPS} \cite{zhang2018unreasonable}: LPIPS evaluates the diversity of the translated images and is demonstrated to correlate well with human perceptual similarity. It is computed as the average LPIPS distance between pairs of randomly sampled translation outputs from the same input. A higher LPIPS score means a more realistic, diverse translated result.
\item \textbf{FCN scores} \cite{isola2017image}: This metric is mostly used in the translation of \textit{semantic maps $\leftrightarrow$real photos} (e.g., c. in Fig.\ref{fig:two-domain}). It uses the FCN-8s architecture \cite{long2015fully} for semantic segmentation to predict a label map for a translated photo and then compares this label map with ground truth labels with standard semantic segmentation metrics, such as per-pixel accuracy, per-class accuracy, and mean class intersection-over-union (class IOU). A higher score indicates a better translated result. 
\item \textbf{Classification accuracy} \cite{liu2017unsupervised}: This metric adapts a classifier pretrained on target domain images to classify the translated images. The intuition behind this metric is that a well-trained I2I model would generate outputs that can be classified as an image from the target domain. A higher accuracy indicates that the model learns more deterministic patterns to be represented in the target domain.
\item \textbf{Density and Coverage (DC)~\cite{naeem2020reliable} is the latest metric for simultaneously judging the diversity and fidelity of generative models. It measures the distance between real images and generated images by introducing a manifold estimation procedure. Higher scores indicate larger diversity and better coverage to the ground-truth domain, respectively.}

\end{itemize}

\begin{table*}
	\centering
	\caption{List of two-domain I2I methods including model name, publication year, the type of training data, whether multimodal or not and corresponding insights.}
	\label{table:list_two}
	\resizebox{\textwidth}{!}{
		\begin{tabular}{c|c|c|c|l}
			\hline
			Method & Publication & Data & Multimodal & \multicolumn{1}{c}{Insights} \\\hline
			pix2pix & 2017 & paired & No & conditional GAN; \\
			DRPAN & 2018 & paired & No & reviser module; \\
			pix2pixHD & 2018 & paired & No & high-resolution; multi-scale architecture; \\
			\cite{AlBahar_2019_ICCV} & 2019 & paired & No & controllable, user-specific generation; \\
			SelectionGAN & 2019 & paired & No & cross-view translation; attention selection; \\
			SPADE & 2019 & paired & No & spatially-adaptive normalization layer; \\
			SEAN & 2020 & paired & No & semantic region-adaptive normalization layer; \\
			CoCosNet & 2020 & paired & No & dense semantic correspondence; \\
			CoCosNetv2 & 2021 & paired & No & PatchMatch; \\
			ASAPNet & 2021 & paired & No & pointwise; non-linear transformation; MLP; \\
			BicycleGAN & 2017 & paired & Yes & cVAE-GAN; cLR-GAN; \\
			PixelNN & 2018 & paired & Yes & nearset-neighbor approach; \\
			\cite{gonzalez2018image} & 2018 & paired & Yes & disentangled representation;\\\hline
			TCR & 2020 & paired+unpaired & No  & transformation consistency regularization;\\\hline
			DTN & 2016 & unpaired & No & invariant representation; domain adaptation; \\
			DualGAN/DiscoGAN/CycleGAN & 2017 & unpaired & No & cyclic loss; \\
			UNIT & 2017 & unpaired & No & cyclic loss; shared latent space; \\
			SCAN & 2018 & unpaired & No & cyclic loss; multi-stage generation; \\
			U-GAT-IT & 2019 & unpaired & No & cyclic loss; cam attention; auxiliary classifier; \\
			GANimorph & 2018 & unpaired & No & cyclic loss; large domain gaps; semantic segmentation; \\
			TraVeLGAN & 2019 & unpaired & No & cyclic loss; large domain gaps; Siamese network; \\
			TransGaGa & 2019 & unpaired & No & cyclic loss; large domain gaps; disentangled representation; \\
			ACL-GAN & 2020 & unpaired & No & large domain gaps; adversarial-consistency loss; \\
			\cite{DBLP:conf/eccv/KatzirLC20} & 2020 & unpaired & No & cyclic loss; large domain gaps; pretrained VGG; cascaded translation; \\
			DistanceGAN & 2017 & unpaired & No & one-sided UI2I; pairwise distances matching; \\
			GCGAN & 2019 & unpaired & No & one-sided UI2I; geometric transformation perservation; \\
			CUT & 2020 & unpaired & No & one-sided UI2I; contrastive learning; \\
			\cite{park2020swapping} & 2020 & unpaired & No & one-sided UI2I; disentanglement; contrastive learning; \\
			TSIT & 2020 & unpaired & No & one-sided UI2I; disentanglement; symmetrical encoders; \\
			F-LSeSim & 2021 & unpaired & No & one-sided UI2I; self-similarity; \\
			LPTN & 2021 & unpaired & No & one-sided UI2I; laplacian pyramid; \\
			DAGAN & 2018 & unpaired & No & instance-level UI2I; attention; \\
			attention-GAN/ & 2018 & unpaired & No & instance-level UI2I; attention; object transfiguration; \\
			attention-guided I2I & 2018 & unpaired & No & instance-level UI2I; attention; cyclic loss; \\
			InstaGAN & 2018 & unpaired & No & instance-level UI2I; segmentation mask; cyclic loss; \\
			INIT\cite{shen2019towards} & 2019 & unpaired & Yes & instance-level UI2I; object+global; cyclic loss; \\
			DUNIT & 2020 & unpaired & Yes & instance-level UI2I; object detector; cyclic loss; \\
			Art2real & 2019 & unpaired & No & segmentation; memory bank; \\
			GDWCT & 2019 & unpaired & No & whitening-and-coloring transformation; \\
			RevGAN & 2019 & unpaired & No & invertible neural networks; \\
			\cite{chen2020distilling} & 2020 & unpaired & No & knowledge distillation; \\
			DAI2I & 2020 & unpaired & No & domain adaptation; \\
			NICE-GAN & 2020 & unpaired & No & introspective adversarial networks; \\
			\cite{kazemi2018unsupervised} & 2018 & unpaired & Yes & domain-specific; domain-invariant; \\
			\cite{pmlr-v80-almahairi18a} & 2018 & unpaired & Yes & augmented CycleGAN; \\
			cd-GAN/MUNIT/DRIT/EGSC-IT & 2018 & unpaired & Yes & disentangled representation; \\
			MSGAN & 2019 & unpaired & Yes & mode-seeking regularization; \\
			\cite{alharbi2019latent} & 2019 & unpaired & Yes & latent filter scaling; \\
			DSMAP & 2020 & unpaired & Yes & domain-specific mapping; \\
			TGAN & 2018 & unpaired & No & few-shot UI2I; transfer learning; \\
			MT-GAN & 2019 & unpaired & No & few-shot UI2I; meta-learning; \\
			EWC & 2020 & unpaired & No & few-shot UI2I; life-long learning; \\
			\cite{ojha2021fewshot} & 2021 & unpaired & No & few-shot UI2I; distance consistency; anchor-based strategy; \\
			OST & 2018 & unpaired & No & one-shot UI2I; sharing layers; selective backpropagation \\
			BiOST & 2019 & unpaired & No & one-shot UI2I; sharing layers; selective backpropagation; bi-direction; \\
			TuiGAN & 2020 & unpaired & No & one-shot UI2I; multi-scale; cyclic loss; \\\hline
	\end{tabular}}
\end{table*}

\section{Two-Domain Image-to-Image Translation}
\label{sec:two-domain}
\begin{figure}[!t]
\centering
\includegraphics[width=0.45\textwidth]{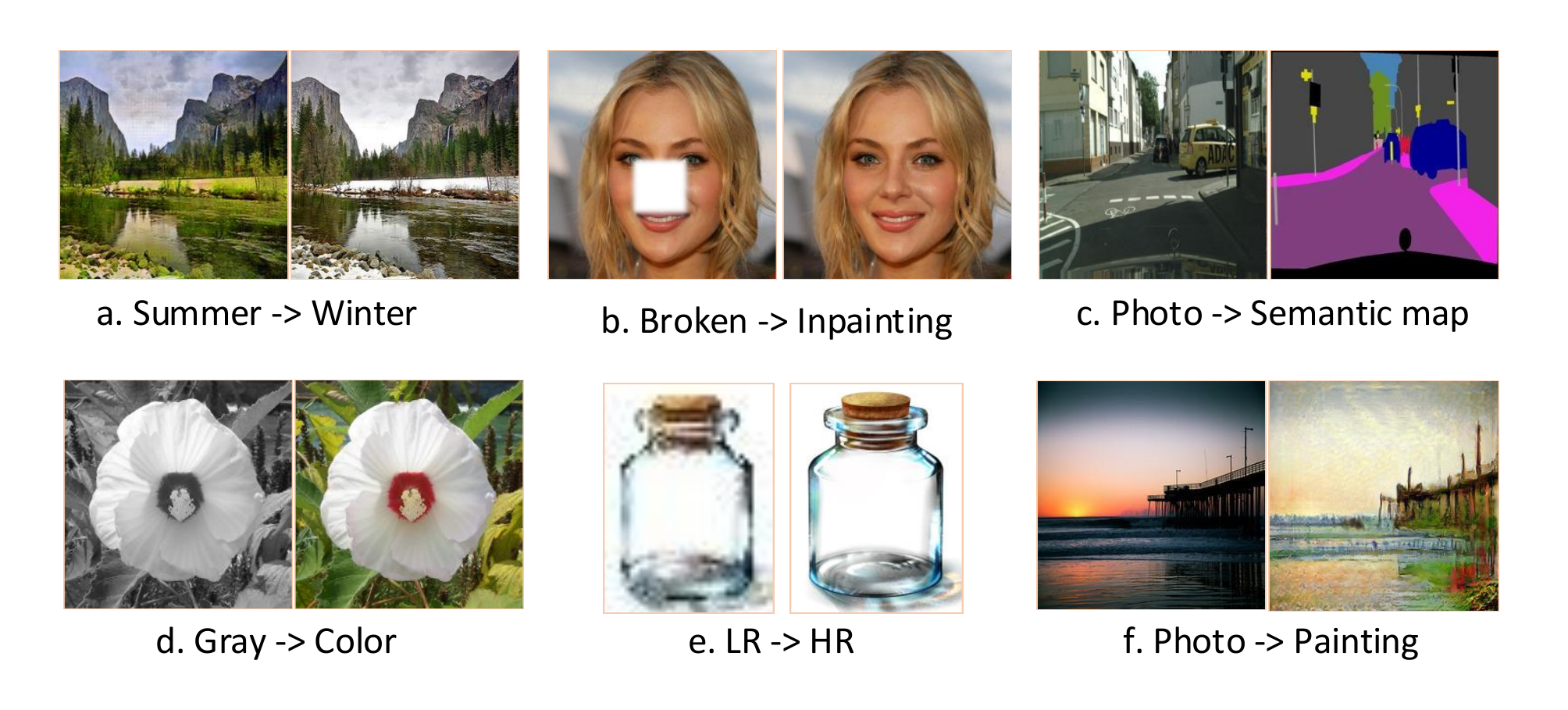}
\caption{Examples of two-domain I2I}
\label{fig:two-domain}
\end{figure}
In this section, we focus on introducing the two-domain I2I methods. As shown in Fig. \ref{fig:two-domain}, two-domain I2I can solve many problems in computer vision, computer graphics and image processing, such as image style transfer (f.) \cite{zhu2017unpaired,lee2020drit++}, which can be used in photo editor apps to promote user experience and semantic segmentation (c.) \cite{Park_2019_CVPR,Zhu_2020_CVPR}, which benefits the autonomous driving and image colorization (d.) \cite{suarez2017infrared,Lee_2020_CVPR}. If low-resolution images are taken as the source domain and high-resolution images are taken as the target domain, we can naturally achieve image super-resolution through I2I (e.) \cite{Yuan_2018_CVPR_Workshops,zhang2019multiple}. 
Indeed, two-domain I2I can be used for many different types of applications as long as the appropriate type and amount of data are provided as the source-target images.
Therefore, we refer to the universal taxonomy in machine learning, such as the categorizations used in \cite{navigli2009word,qi2019small,schmarje2020survey}, and classify two-domain I2I methods into four categories based on the different ways of leveraging various sources of information: supervised I2I, unsupervised I2I, semi-supervised I2I and few-shot I2I, as described in following paragraph. We also provide the summary of these two-domain I2I methods in Table~\ref{table:list_two} including method name, publication year, the type of training data, whether multi-modal or not and corresponding insights.

\begin{itemize}
\item \textbf{Supervised I2I}
In the earlier I2I works \cite{isola2017image}, researchers used many aligned image pairs as the source domain and target domain to obtain the translation model that translates the source images to the desired target images. 
\item \textbf{Unsupervised I2I}
Training supervised translation is not very practical because of the difficulty and high cost of acquiring these large, paired training data in many tasks. Taking photo-to-painting translation as an example (e.g., f. in Fig.\ref{fig:two-domain}), it is almost impossible to collect massive amounts of labeled paintings that match the input landscapes. Hence, unsupervised methods \cite{zhu2017unpaired,kim2017learning,Yi_2017_ICCV} have gradually attracted more attention. In an unsupervised learning setting, I2I methods use two large but unpaired sets of training images to convert images between representations. 
\item \textbf{Semi-supervised I2I}
In some special scenarios, we still need a little expensive human labeling or expert guidance, as well as abundant unlabeled data, such as those of old movie restoration \cite{mustafa2020transformation} or genomics \cite{shi2011semi}. Therefore, researchers consider introducing semi-supervised learning \cite{kingma2014semi,rasmus2015semi,berthelot2019mixmatch} into I2I to further promote the performance of image translation. Semi-supervised I2I approaches leverage only source images alongside a few source-target aligned image pairs for training but can achieve more promoted translated results than their unsupervised counterpart.
\item \textbf{Few-shot I2I}
Nonetheless, several problems remain regarding translation using a supervised, unsupervised or semi-supervised I2I method with extremely limited data. In contrast, humans can learn from only one or limited exemplars to achieve remarkable learning results. As noted by meta-learning \cite{zhang2018metagan,sun2019meta} and few-shot learning \cite{snell2017prototypical,sung2018learning}, humans can effectively use prior experiences and knowledge when learning new tasks, while artificial learners usually severely overfit without the necessary prior knowledge. Inspired by the human learning strategy, few- and one-shot I2I algorithms \cite{Liu_2019_ICCV,lin2020tuigan,lin2019learning,lin2019zstgan} have been proposed to translate from very few (or even one) in the limit unpaired training examples of the source and target domains.
\end{itemize}

\begin{figure}[!t]
\centering
\includegraphics[width=0.45\textwidth]{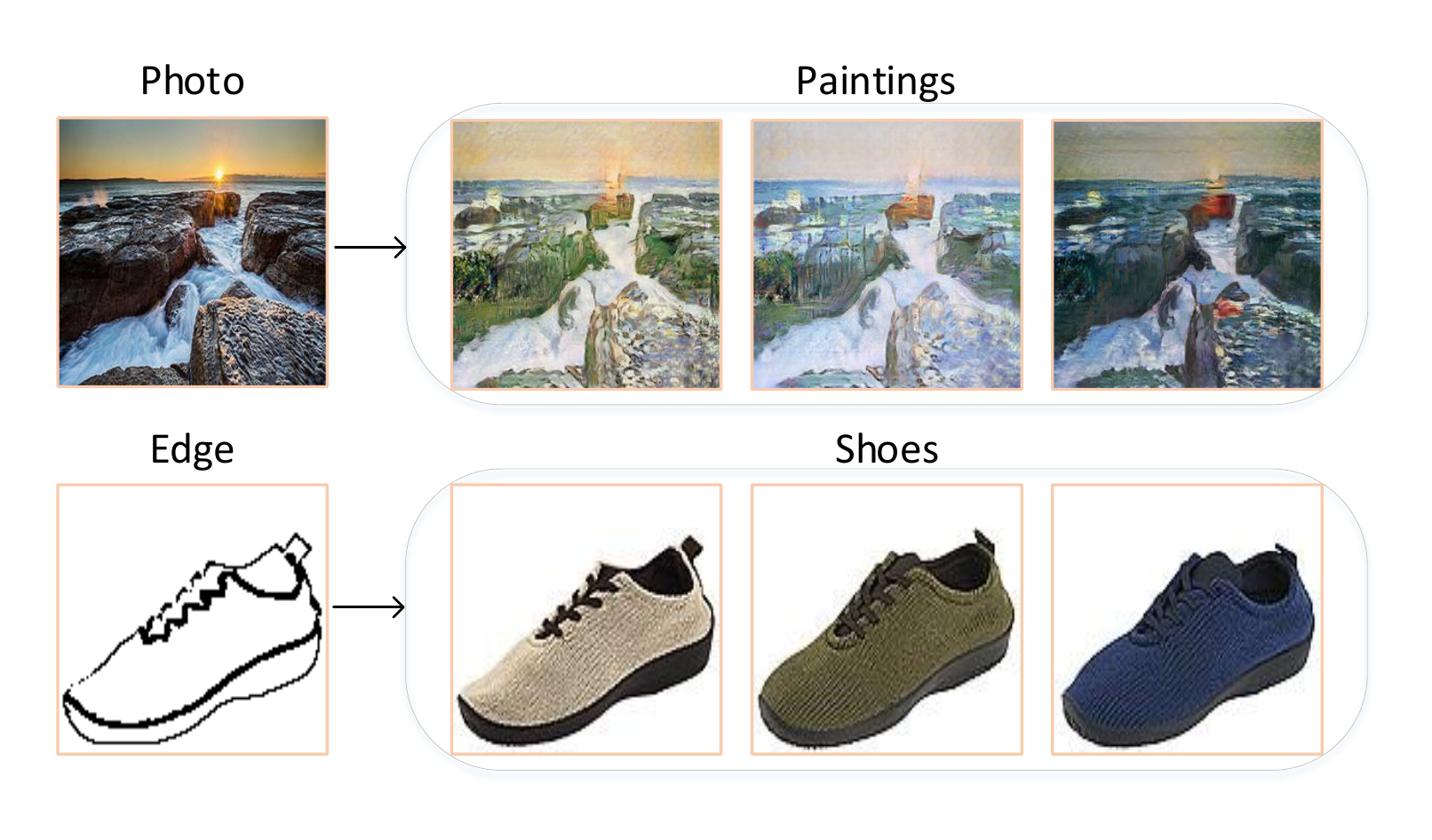}
\caption{Examples of multimodal outputs in two-domain I2I.}
\label{fig:two-domain-multi-dodal}
\end{figure}

Although learning settings may differ, most of these I2I techniques tend to learn a deterministic one-to-one mapping and only generate single-modal output, as shown in Fig.\ref{fig:two-domain}. However, in practice, the two-domain I2I is inherently ambiguous, as one input image may correspond to multiple possible outputs, namely, multimodal outputs, as shown in Fig.\ref{fig:two-domain-multi-dodal}. Multimodal I2I translates the input image from one domain to a distribution of potential outputs in the target domain while remaining faithful to the input. 
These diverse outputs represent different color or style texture themes (i.e., multimodal) but still preserve the similar semantic content as the input source image. Therefore, we actually view multimodal I2I as a special two-domain I2I and discuss it in supervised (subsection \ref{subsec:supervised I2I}) and unsupervised settings (subsection \ref{subsec:UI2I}).

\subsection{Supervised Image-to-Image Translation}
\label{subsec:supervised I2I}
Supervised I2I aims to translate source images into the target domain with many aligned image pairs as the source domain and target domain for training. In this subsection, we further divide the supervised I2I in two categories: methods with single-modal output and methods with multimodal outputs. 
\subsubsection{Single-modal Output} 
The idea of I2I can be traced back to Hertzmann et al.’s image analogies \cite{hertzmann2001image}, which use a non-parametric texture model for a wide variety of “image filter” effects with an image pair input. More recent research on I2I mainly leverages the deep convolutional neural network to learn the mapping function. Isola et al. \cite{isola2017image} first apply conditional GAN to an I2I problem by proposing pix2pix to solve a wide range of supervised I2I tasks. In addition to the pixelwise regression loss $\mathcal{L}_{1}$ between the translated image and the ground truth, pix2pix leverages adversarial training loss $\mathcal{L}_{cGAN}$ to ensure that the outputs cannot be distinguished from “real” images. The objective is:
\begin{equation}
\mathcal{L} = \mathop{min}\limits_{G}\mathop{max}\limits_{D}\mathcal{L}_{cGAN}(G,D) +\lambda\mathcal{L}_{\mathcal{L}_{1}}(G).
\label{eq:pix2pix}
\end{equation}
Pix2pix is also a strong baseline image translation framework that inspires many improved I2I works based on it, as described in following parts. 

Wang et al. \cite{Wang_2018_ECCV} claim that the GAN loss and pixelwise loss used in pix2pix often lead to blurry results. They present discriminative region proposal adversarial networks (DRPANs) to address it by adding a reviser ($R$) to distinguish real from masked fake samples. 
Wang et al. \cite{wang2018high} argue that the adversarial training in pix2pix \cite{isola2017image} might be unstable and prone to failure for high-resolution image generation tasks. They propose an HD version of pix2pix that can increase the photo realism and resolution of the results to 2048×1024. 
Moreover, AlBahar et al. \cite{AlBahar_2019_ICCV} take an important step toward addressing the controllable or user-specific generation based on pix2pix \cite{isola2017image} via respecting the constraints provided by an external, user-provided guidance image.

Unfortunately, pix2pix \cite{isola2017image} and its improved variants \cite{Wang_2018_ECCV,wang2018high,AlBahar_2019_ICCV} still fail to capture the complex scene structural relationships through a single translation network when the two domains have drastically different views and severe deformations. Tang et al. \cite{tang2019multi} therefore proposed SelectionGAN to solve the cross-view translation problem, i.e., translating source view images to target view scenes in which the fields of views have little or no overlap. It was the first attempt to combine the multichannel attention selection module with GAN to solve the I2I problem. 

What's more, SPADE \cite{Park_2019_CVPR} proposes the spatially-adaptive normalization layer to further improve the quality of the synthesized images. But SPADE uses only one style code to control the entire style of an image and inserts style information only in the beginning of a network. SEAN \cite{Zhu_2020_CVPR} therefore designs semantic region-adaptive normalization layer to alleviate the two shortcomings.

Having said that, Shaham et al.~\cite{shaham2020spatiallyadaptive} claim that traditional I2I networks~\cite{isola2017image,wang2018high,Park_2019_CVPR} suffer from acute computational cost when operating on high-resolution images. They propose to design a more lightweight but efficient enough network ASAPNet for fast high-resolution I2I.

Recently, Zhang et al. \cite{zhang2020cross} proposed an exemplar-based I2I framework, CoCosNet, to translate images by establishing the dense semantic correspondence between cross-domain images. However, the semantic matching process may lead to a prohibitive memory footprint when estimating a high-resolution correspondence. Zhou et al. \cite{zhou2020full} therefore proposed a GRU-assisted refinement module that applies PatchMatch in a hierarchy to first learn the full-resolution, 1024×1024, cross-domain semantic correspondence, namely CoCosNetv2.

\subsubsection{Multimodal Outputs} 
As shown in Fig.\ref{fig:two-domain-multi-dodal}, multimodal I2I translates the input image from one domain to a distribution of potential outputs in the target domain while remaining faithful to the input. 

Actually, this multimodal translation benefits from the solutions of \textit{mode collapse problem} \cite{goodfellow2017nips,arjovsky2017wasserstein,gulrajani2017improved}, in which the generator tends to learn to map different input samples to the same output. Thus, many multimodal I2I methods \cite{zhu2017toward,bansal2018pixelnn} focus on solving the mode collapse problem to lead to diverse outputs naturally. BicycleGAN \cite{zhu2017toward} became the first supervised multimodal I2I work by combining cVAE-GAN \cite{hinton2006reducing,kingma2014auto,larsen2016autoencoding} and cLR-GAN \cite{chen2016infogan,donahue2016adversarial,dumoulin2016adversarially} to systematically study a family of solutions to the mode collapse problem and generate diverse and realistic outputs.

Similarly, Bansal et al. \cite{bansal2018pixelnn} proposed PixelNN to achieve multimodal and controllable translated results in I2I. They proposed a nearest-neighbor (NN) approach combining pixelwise matching to translate the incomplete, conditioned input to multiple outputs and allow a user to control the translation through on-the-fly editing of the exemplar set. 

Another solution for producing diverse outputs is to use \textit{disentangled representation} \cite{chen2016infogan,Higgins2017betaVAELB,kim2018disentangling,NIPS2017_7028} which aims to break down, or disentangle, each feature into narrowly defined variables and encodes them as separate dimensions. When combining it with I2I, researchers disentangle the representation of the source and target domains into two parts: domain-invariant features \textit{content}, which are preserved during the translation, and domain-specific features \textit{style}, which are changed during the translation. In other words, I2I aims to transfer images from the source domain to the target domain by preserving \textit{content} while replacing \textit{style}. Therefore, one can achieve multimodal outputs by randomly choosing the \textit{style} features that are often regularized to be drawn from a prior Gaussian distribution $N(0, 1)$. Gonzalez-Garcia et al. \cite{gonzalez2018image} disentangled the representation of two domains into three parts: the \textit{shared} part containing common information of both domains, and two \textit{exclusive} parts that only represent those factors of variation that are particular to each domain. In addition to the bi-directional multimodal translation and retrieval of similar images across domains, they can also transfer a domain-specific transfer and interpolation across two domains.

\subsection{Unsupervised Image-to-Image Translation (UI2I)}
\label{subsec:UI2I}
UI2I uses two large but unpaired sets of training images to convert images from one representation to another. In this subsection, we follow the same categories in subsection \ref{subsec:supervised I2I}: single-modal output and multimodal outputs.

\subsubsection{Single-modal Output}
\begin{figure}[!t]
\centering
\includegraphics[width=0.45\textwidth]{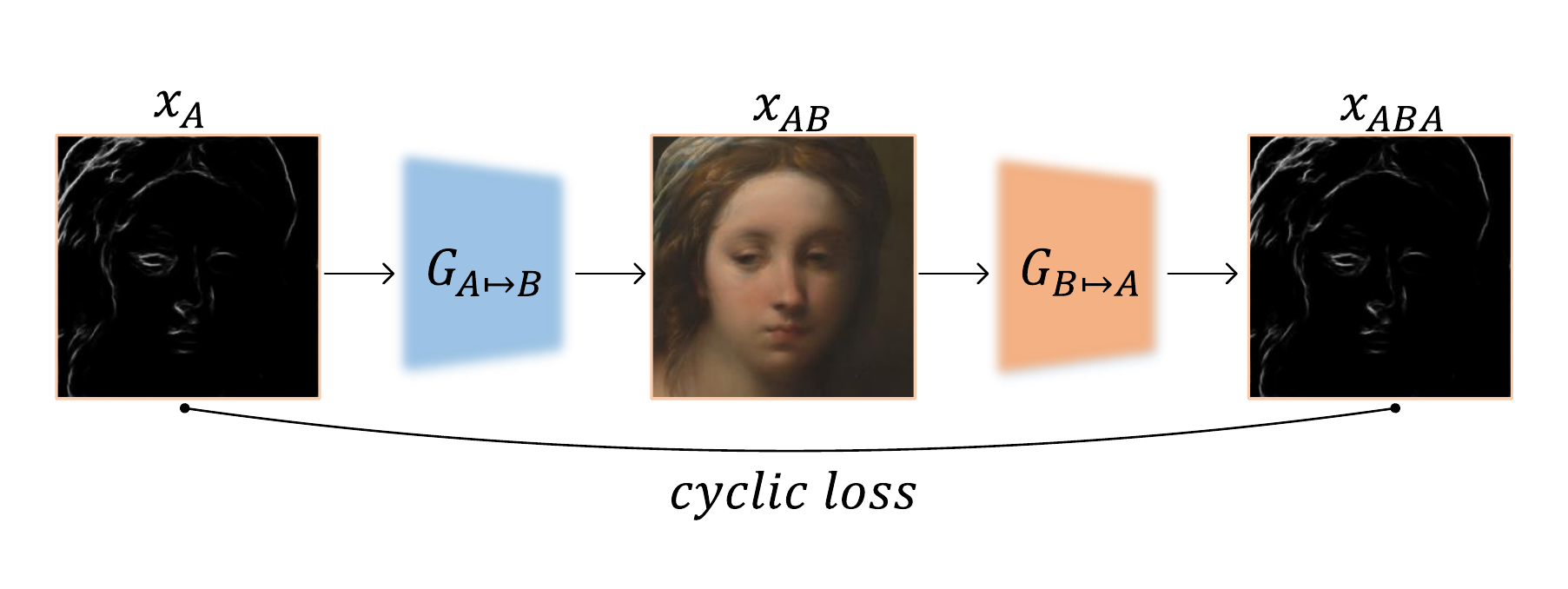}
\caption{Taking edge $\mapsto$ face translation as an example, we use a cycle-consistency constraint between a source image $x_{A}$ and its cyclic reconstructed image $x_{ABA}$, termed cyclic loss through two translators $G_{A\mapsto{B}}$ and $G_{B\mapsto{A}}$.}
\label{fig:cyclic_loss}
\end{figure}

\begin{figure}[!t]
\centering
\includegraphics[width=0.45\textwidth]{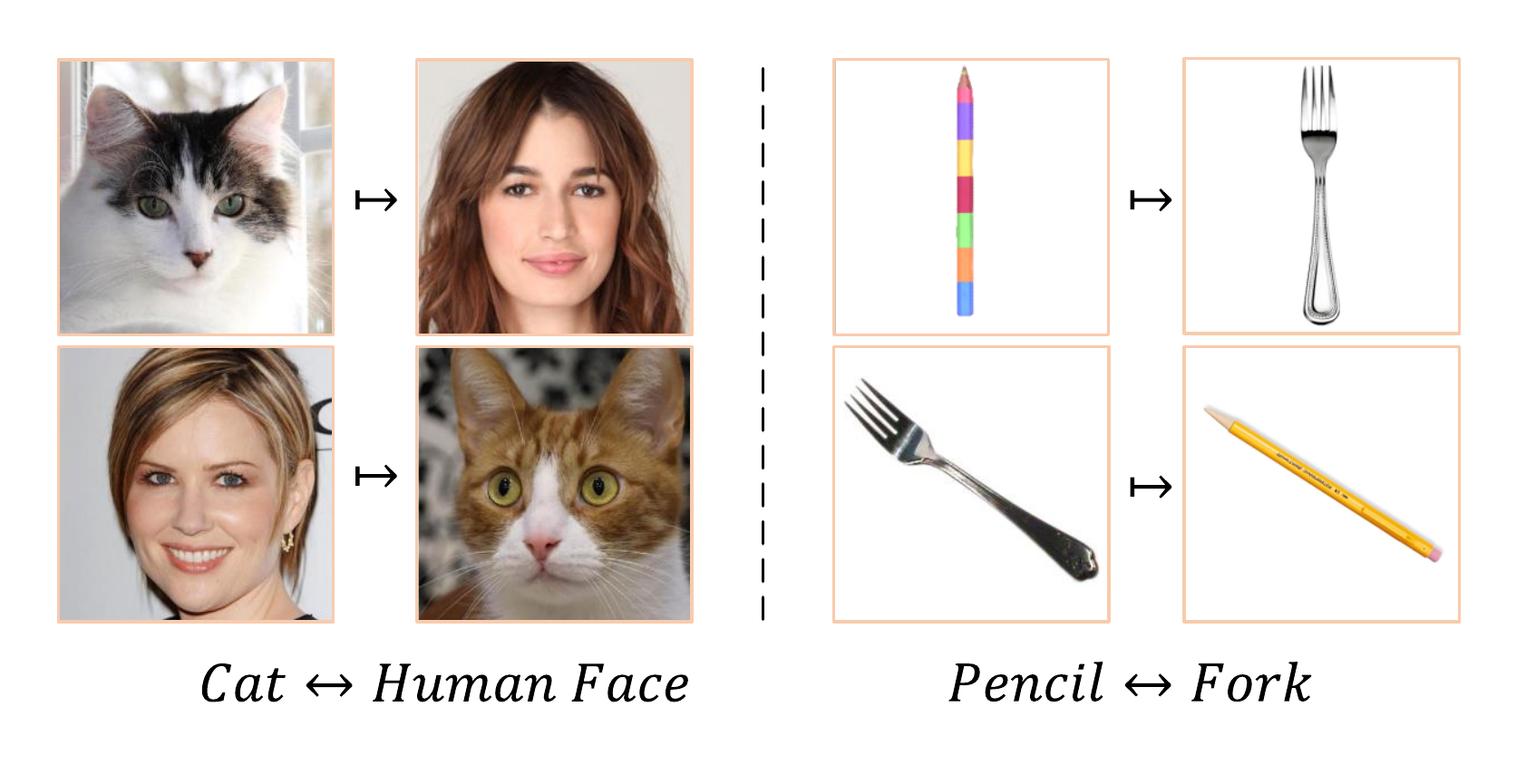}
\caption{Examples of I2I with large domain gaps, where the translated images are sufficiently realistic in the target domain and preserve semantic information learned from the source domain.}
\label{fig:large_domain_gaps}
\end{figure}

\begin{figure}[!t]
\centering
\includegraphics[width=0.5\textwidth]{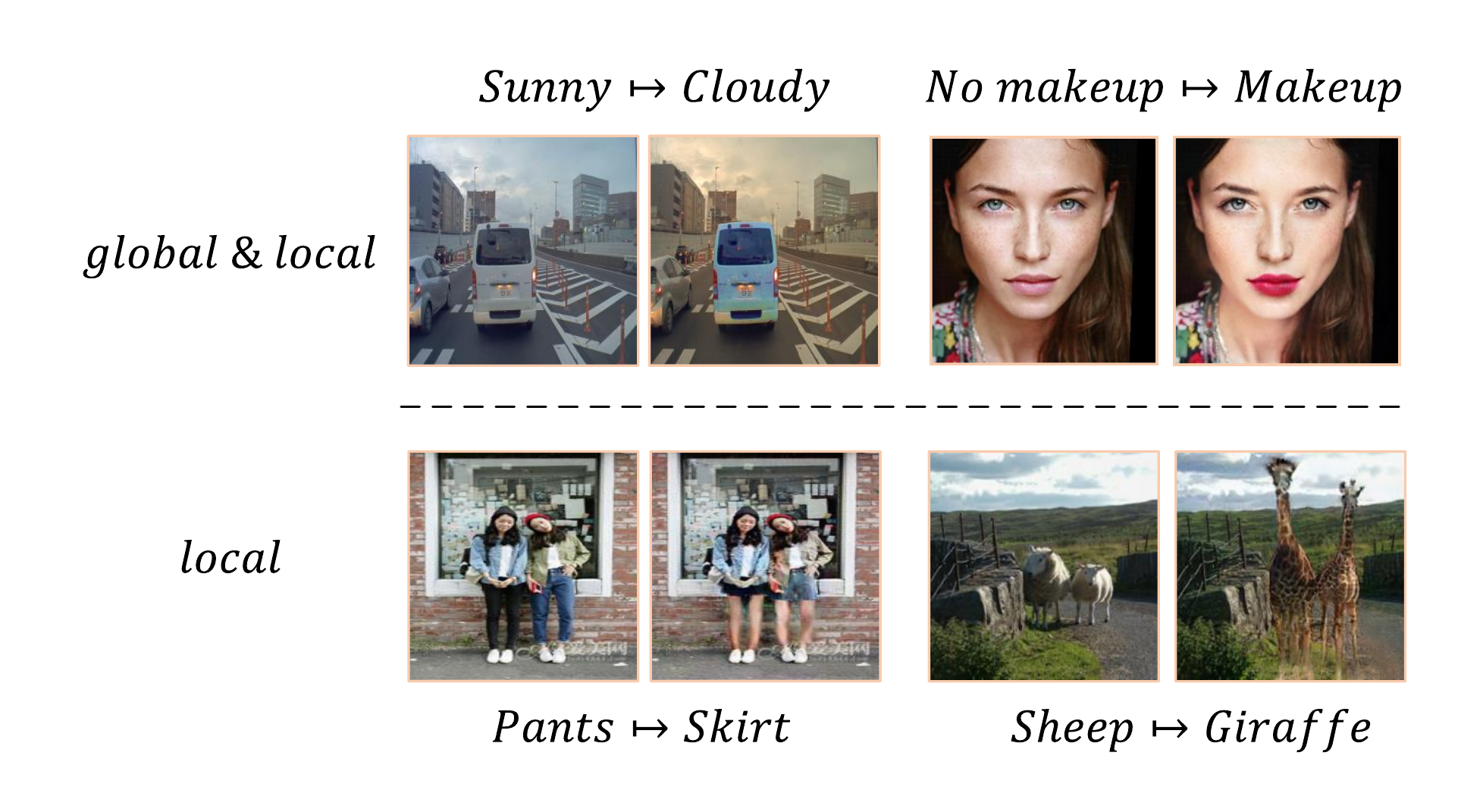}
\caption{Examples of I2I focusing on fine-grained objects. The top examples generate results with the global style (weather or foundation) translated and the local object instance (car or lipstick) changed. The bottom examples achieve remarkable translated results for fine-grained local objects.}
\label{fig:fine_grained_obj}
\end{figure}

UI2I methods have been explored primarily by focusing on different issues. We will introduce those methods with single-modal output in the following four categories: translation using a cycle-consistency constraint, translation beyond a cycle-consistency constraint, translation of fine-grained objects and translation by combining knowledge in other fields. 
\begin{itemize}
\item \textbf{Translation using a Cycle-consistency Constraint} In the beginning, researchers tried to find new frameworks or constraints to establish the I2I mapping without labels or pairings. Based on this motivation, the cycle-consistency constraint was proposed, as shown in Fig.\ref{fig:cyclic_loss} and was proved to be an effective strategy for overcoming the lack of supervised pairing. 
\item \textbf{Translation beyond Cycle-consistency Constraint} However, while the cycle-consistency constraint can eliminate the dependence on supervised paired data, it tends to force the model to generate a translated image that contains all the information of the input image for reconstructing the input image. Approaches using cyclic loss are typically unsuccessful when the two domains require substantial clutter and heterogeneity instead of small, simple changes in low-level shape and context. Therefore, many UI2I methods focus on the translation beyond the cycle-consistency constraint, as shown in Fig.\ref{fig:large_domain_gaps}, to solve the homogeneous limitation, as well as the large shape deformation problem between the source and target domains. 
\item \textbf{Translation of Fine-grained Objects} Most UI2I models using or beyond the cycle-consistency constraint tend to directly synthesis a new domain with the global target style translated and give little thought of the local objects or fine-grained instances during translation. However, in some application scenarios, such as virtual try-on, we may only need to change a local object, such as changing pants to a skirt with other parts unchanged, as shown in Fig.\ref{fig:fine_grained_obj}. In this case, severe setbacks are incurred when the translation involves large shape changes of instances or multiple discrepant objects. Hence, research on applying instances or objects information in UI2I is a growing trend.
\item \textbf{Translation by combining knowledge in other fields} In addition, some UI2Is try to improve the network efficiency or translation performance by combining knowledge from other research areas. 
\end{itemize}

\paragraph{Translation using a Cycle-consistency Constraint}
The popularly known strategy for tackling an unsupervised setting is to use the cycle-consistency constraint (cyclic loss) shown in Fig. \ref{fig:cyclic_loss}. Cyclic loss uses two translators, $G_{A\mapsto{B}}$ and $G_{B\mapsto{A}}$, to define a cycle-consistency loss between the source image $x_{A}$ and its reconstruction $x_{ABA}$ when the pairs are not available, and the objective can be written as:
\begin{equation}
\mathcal{L}_{cyc}=\mathcal{L}(x_{A},G_{B\mapsto{A}}(G_{A\mapsto{B}}(x_{A}))).
\end{equation}

Taigman et al. \cite{DBLP:conf/iclr/TaigmanPW17} present a domain transfer network (DTN) for unpaired cross-domain image generation by assuming constant latent space between two domains, which could generate images of the target domains' style and preserve their identity. 
Similar to the idea of dual learning in neural machine translation, DualGAN \cite{Yi_2017_ICCV}, DiscoGAN \cite{kim2017learning} and CycleGAN \cite{zhu2017unpaired} are proposed to train two cross-domain transfer GANs with two cyclic losses at the same time. 
Liu et al. \cite{liu2017unsupervised} propose UNIT to make a shared latent space assumption that a pair of corresponding images in different domains can be mapped to the same latent code in a shared latent space. They show that the shared-latent space constraint implies the cycle-consistency constraint.
Li et al. \cite{li2018unsupervised} claim that these single-stage unsupervised approaches are difficult to use for translating two-domain images with high-resolution or a substantial visual gap. They hence propose a stacked cycle-consistent adversarial network (SCAN) to decompose the single complex image translation process into multistage transformations. 
More recently, Kim et al. \cite{kim2019u} proposed U-GAT-IT to incorporate a novel attention module to force the generator and discriminator to focus on more important regions via the auxiliary classifier.

\paragraph{Translation beyond Cycle-consistency Constraint} To address the challenging shape deformation problem (i.e., large domain gaps) in I2I shown in Fig. \ref{fig:large_domain_gaps}, Gokaslan et al. \cite{gokaslan2018improving} propose GANimorph to reframe the discrimination problem from determining real or fake images into a semantic segmentation task of finding real or fake regions of the image with dilated convolutions. 
Amodio et al. \cite{amodio2019travelgan} introduce TraVeLGAN to address the challenge. In addition to the generator and discriminator, they add a Siamese network to define a transformation vector between two images of each domain and minimize the distance between the two vectors. The Siamese network guides the generator such that each original image shares semantics with its generated version.
Wu et al. \cite{wu2019transgaga} present TransGaGa to solve the large geometry variations in I2I. They disentangle each domain into a Cartesian product of the geometry space and appearance space by the VAE. In each space, they apply a bi-direction geometry transformation and appearance transformation with two transformers, respectively. 
Zhao et al. \cite{zhao2020unpaired} argue that I2I can barely perform shape changes, remove objects or ignore irrelevant texture because of the strict pixel-level constraint of cycle-consistent loss. They propose ACL-GAN, which uses a novel adversarial-consistency loss to replace the cyclic loss to maintain the commonalities across two domains. 
Recently, Katzir et al. \cite{DBLP:conf/eccv/KatzirLC20} mitigated shape translation in a cascaded, deep-to-shallow fashion, in which they exploited the deep features extracted from a pretrained VGG-19 and translated them at the feature level.

Moreover, some UI2I works try to design a one-side translation process to remove the cycle-consistency constraint. 
These methods usually take into account some kind of geometry distance as content loss between the original source image and translated results.  Benaim et al.~\cite{benaim2017one} propose DistanceGAN to achieve one-side translation by maintaining the distances between images within domains. Fu et al.~\cite{fu2019geometry} propose GCGAN to preserve the given geometric transformation between the input images before and after translation.
Zheng et al.~\cite{zheng2021spatiallycorrelative} propose F-LSeSim to learn a domain-invariant representation to precisely express scene structure via self-similarity.
Liang et al.~\cite{liang2021high} propose a Laplacian Pyramid Translation Network (LPTN) to achieve photorealistic I2I by decomposing the input into a Laplacian pyramid and translating on the low-frequency component.
Park et al. \cite{park2020contrastive} propose CUT to maximize the mutual information between the input-output pairs via contrastive learning \cite{oord2018representation} in a patch-based way rather than operating on entire images. 
Jiang et al. \cite{jiang2020tsit} propose a symmetrical two-stream framework (TSIT) to learn feature-level semantic structure information and style representation, and then they exploit the generator to fuse content and style feature maps from coarse to fine. 
Park et al. \cite{park2020swapping} also propose a swapping autoencoder for texture swapping by enforcing the output and reference patches to appear indistinguishable via the patch co-occurrence discriminator.


\paragraph{Translation of Fine-grained Objects}
Some I2I works try to translate on a higher semantic level by replacing the local texture information of object instances as well as the global style translated, as shown in Fig.\ref{fig:fine_grained_obj}.
Ma et al. \cite{ma2018gan} propose DAGAN to construct a deep attention encoder to enable the instance-level translation. 
Chen et al. \cite{Chen_2018_ECCV} and Mejjati \cite{mejjati2018unsupervised} almost simultaneously proposed attention GAN and attention-guided I2I, respectively, focusing on achieving an I2I translation of individual objects without altering the background, as shown in Fig.\ref{fig:fine_grained_obj} (local). 
Mo et al. \cite{mo2018instagan} propose InstaGAN, which is the first work to solve multi-instance transfiguration tasks in UI2I. It uses the object segmentation masks to translate an image and the corresponding set of instance attributes while maintaining the permutation invariance property of instances. Shen et al. \cite{shen2019towards} propose the instance-aware I2I approach (INIT) to use the fine-grained local instances based on MUNIT \cite{huang2018multimodal} and DRIT \cite{lee2018diverse}. DUNIT \cite{Bhattacharjee_2020_CVPR} incorporates an object detector within the I2I architecture used in DRIT\cite{lee2018diverse} to leverage the object instances to reassemble the resulting representation. 

\paragraph{Translation by Combining Knowledge in Other Fields}
Tomei et al. \cite{tomei2019art2real} present Art2Real to translate artistic paintings to real photos using a weakly supervised segmentation model and memory banks.
Inspired by the style transfer, Cho et al. \cite{cho2019image} propose GDWCT, which extends whitening-and-coloring transformation (WCT) to I2I to achieve a highly competitive image quality. Chen et al. \cite{chen2020distilling} use the knowledge distillation scheme to define a teacher generator and student discriminator. A distilling portable model is shown to achieve a comparable performance with substantially lower memory usage and computational cost. With the help of domain adaptation, Chen et al. \cite{Chen_2020_CVPR} develop DAI2I to adapt a given I2I model trained on the source domain to a new domain, which improves the generalization capacity of existing models. RevGAN \cite{van2019reversible} interpolates the invertible neural networks (INNs) into I2I to reduce memory overhead, as well as increase the fidelity of the output. NICE-GAN \cite{chen2020reusing} first reuses the discriminator for embedding from images to hidden vectors (as encoding) in which the discriminator is conducted using introspective adversarial networks (IANs). It derives a more compact and effective architecture for generating translated images.

\subsubsection{Multi-modal Outputs}
Kazemi et al. \cite{kazemi2018unsupervised} show that shared-latent space assumptions only model the domain-invariant information across two domains and fail to capture the domain-specific information. They argue for learning a one-to-many UI2I mapping by extending CycleGAN to learn a domain-specific code for each domain jointly with a domain-invariant code.
Similarly, Almahairi et al. \cite{pmlr-v80-almahairi18a} also claim that the mapping across two domains should be characterized as many-to-many instead of one-to-one. They therefore introduce the augmented CycleGAN model to capture the diversity of the outputs, i.e., multimodal, by extending CycleGAN's training procedure to the augmented spaces.

\begin{figure}[!t]
\centering
\includegraphics[width=0.45\textwidth]{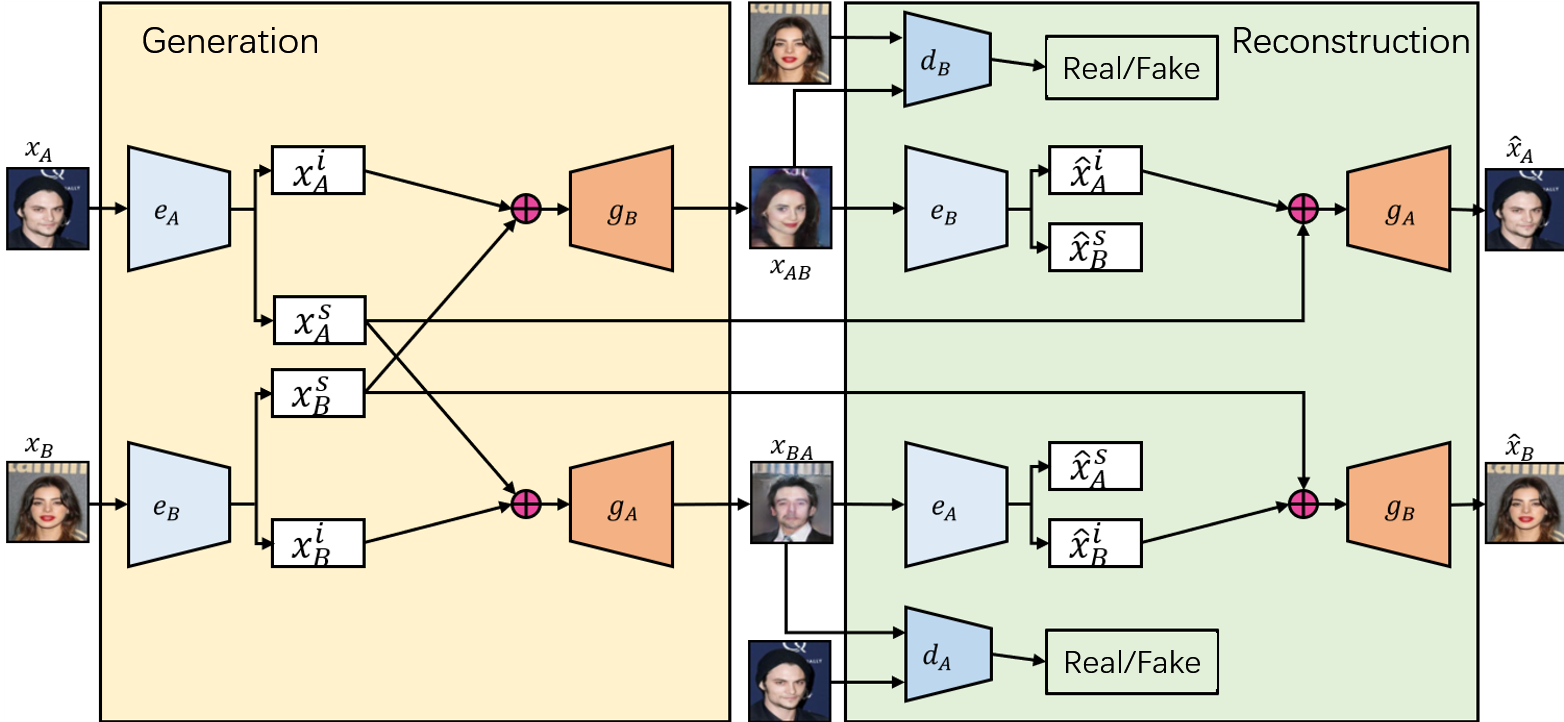}
\caption{The architecture of cd-GAN \cite{Lin_2018_CVPR}.}
\label{fig:cd-gan}
\end{figure}
Disentangled representations \cite{chen2016infogan,Higgins2017betaVAELB,kim2018disentangling} also offer a solution to this problem in unsupervised settings. They inspire multimodal UI2I advances, such as cd-GAN \cite{Lin_2018_CVPR}, MUNIT \cite{huang2018multimodal}, DRIT \cite{lee2018diverse} and EGSC-IT \cite{ma2018exemplar}, which were proposed almost simultaneously and present similar model designings, such as that of Fig.\ref{fig:cd-gan}. For example, DRIT \cite{lee2018diverse} assumes embedding images onto two spaces, domain-invariant content space and domain-specific attribute space, via a content encoder and attribute encoder. Specifically, DRIT learns two objective losses, content adversarial loss and cross-cycle consistency loss. Through content adversarial loss, it applies weight sharing and a content discriminator to force content representation to be mapped onto the same shared space, as well as to guarantee the same content representations encode the same information for both domains. Then, with the constraint of cross-cycle consistency loss, it performs forward-backward translation by swapping domain-specific representations. At test time, DRIT can use different attribute vectors randomly sampled from domain-specific attribute space to generated diverse outputs.

However, these aforementioned methods still cannot solve the problem of target domain images that are content-rich with multiple discrepant objects. Shen et al. \cite{shen2019towards} therefore propose INIT to translate instance-level objects and background/global areas separately with different style codes. Chang et al. \cite{chang2020domainspecific} declare that the shared domain-invariant content space in disentangled representations could limit the ability to represent content because these representations ignore the relationship between content and style. They present DSMAP to leverage two extra domain-specific mappings to remap the content features from shared domain-invariant content space to two independent domain-specific content spaces for two domains.

Other attempts to address multimodal UI2I are proposed by Mao et al. \cite{mao2019mode} and Alharbi et al. \cite{alharbi2019latent}. The study in \cite{mao2019mode} uses MSGAN to employ a mode-seeking regularization method to solve the mode collapse problem in cGANs, and the proposed regularization method can be readily integrated with an existing cGANs framework, such as DRIT \cite{lee2018diverse}, to generate more diverse translated images. In addition, \cite{alharbi2019latent} uses latent filter scaling (LFS) to perform multimodal UI2I, which is the first multimodal UI2I framework that does not require autoencoding or reconstruction losses for the latent codes or images.

\subsection{Semi-Supervised Image-to-Image Translation}
Semi-supervised I2I draws much attention in some special applications, such as old movie restoration or artistic reconstructions. In these scenarios, one needs few human-labeled data for guidance and abundant, unlabeled other data for automatic translation. Unpaired data, when used in conjunction with a small amount of paired data, can produce considerable improvement in translation performance. 

Mustafa et al. \cite{mustafa2020transformation} first study the applicability of semi-supervised learning in a two-domain I2I setting.
They introduce a regularization term, transformation consistency regularization (TCR), to force a model’s prediction to remain unchanged for the perturbed (geometric transform) input sample and its reconstruction version. In detail, they train an I2I mapping model $f_{\theta}$ by minimizing the supervised loss $\mathcal{L}_{s}$ with paired source-target images $(x_{i},y_{i})$:
\begin{equation}
\mathcal{L}_{s}=mse(y_i, f_{\theta}(x_{i})).
\end{equation}
Then, they leverage unsupervised data to regularize the model’s predictions over varied forms of geometric transformations $T_{m}$. They make use of $T_{m}$ to process the unlabeled input samples and feed these transformed samples into the I2I model $f_{\theta}$ to obtain the disturbed outputs $f_{\theta}(T_{m}(u_{i}))$. On the other hand, they directly feed the unlabeled samples into $f_{\theta}$ to acquire primary outputs and apply geometric transformations $T_{m}$ onto them to obtain another type of perturbed outputs ${T_{m}(f_{\theta}(u_{i})}$. The TCR of unlabeled data guarantees the consistency between the two outputs to learn more about the inherent structure of the source and target domain distributions. The detailed unsupervised TCR regularization loss $\mathcal{L}_{us}$ is as follows:
\begin{equation}
\mathcal{L}_{us}=mse({T_{m}(f_{\theta}(u_{i}),f_{\theta}(T_{m}(u_{i})))}).
\end{equation}
Their method can use unlabeled data and less than $1\%$ of labeled data to complete several I2I tasks, such as image colorization, image denoising and image super-resolution.

\subsection{Few-Shot Image-to-Image Translation}
Existing I2I models cannot translate images from very few (even one) training examples of the source and target domains. In contrast, humans can learn from very limited exemplars to obtain extraordinary learning results. For example, a child can recognize what a "zebra" and "rhino" are with only a few pictures. Inspired by the rapid learning ability of humans, researchers expect that after the machine learning model has learned a large amount of data in a certain category, for new categories, it only needs a few samples to learn quickly. In other words, few-shot I2I wants to solve the transfer capability or generalization ability of the I2I model given very few samples. 

Drawing inspiration from domain adaptation or transfer learning, some methods solve the few-shot translation by adapting a pretrained-network trained on large-scale source domain to the target domain with a few images only. Wang et al.~\cite{wang2018transferring} propose Transferring GAN (TGAN) to successful combine transfer learning with GAN. It transfers images from source domain to target domain with a pretrained network when data is limited. Lin et al. propose MT-GAN \cite{lin2019learning} to incorporate prior from previous domain translation tasks when assuming a new domain translation task from a perspective of meta-learning. Likewise, Li et al.~\cite{li2020few} propose to utilize EWC to adapt the weights of pretrained network on source domain to a new target domain. Ojha et al.~\cite{ojha2021fewshot} achieve few-shot adaptation by a novel cross-domain distance consistency loss and an anchor-based strategy.   


An extreme scenario of few-shot I2I is one-shot I2I. Benaim et al.\cite{benaim2018one} propose OST to solve the one-shot cross-domain translation problem, which aims to learn a unidirectional mapping function given a single source image and a set of images from the target domain. By sharing layers between the two autoencoders and selective backpropagation, OST enforces the same structure on the encoding of both domains, which benefits the translation. 
As an extension of OST, Cohen et al. \cite{cohen2019bidirectional} propose BiOST to translate in bi-direction without a weight sharing strategy via a feature-cycle consistency term. 

\begin{figure}[!t]
\centering
\includegraphics[width=0.45\textwidth]{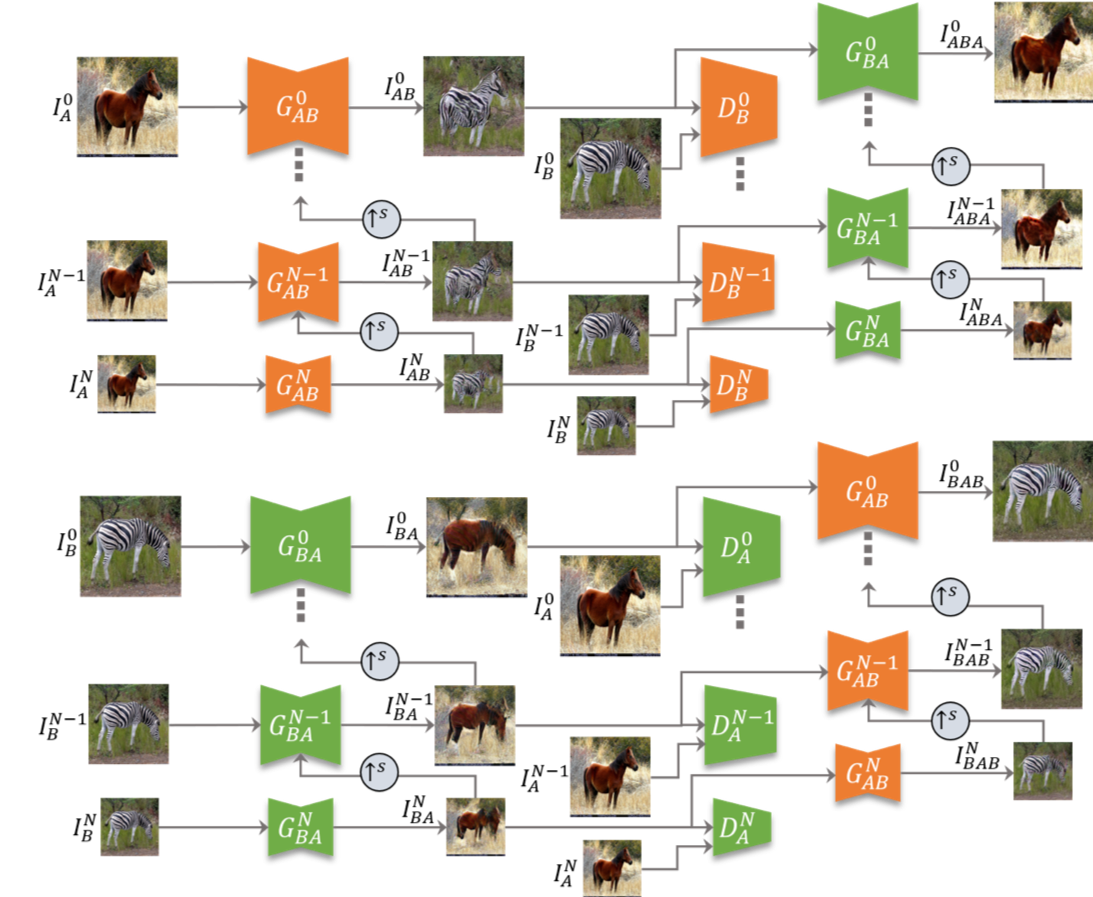}
\caption{The architecture of TuiGAN \cite{lin2020tuigan}.}
\label{fig:tuigan}
\end{figure}
In contrast to the one-shot setting in \cite{benaim2018one,cohen2019bidirectional} that uses a single image from the source domain and a set of images from the target domain, Lin et al. propose TuiGAN \cite{lin2020tuigan} to achieve one-shot UI2I with only two unpaired images from two domains. They train TuiGAN in a coarse-to-fine manner using the cycle-consistency constraint shown in Fig.\ref{fig:tuigan}. In detail, they design two pyramids of generators and discriminators to transfer the domain distribution of the input image to the target domain by progressively translating the image from coarse to fine. Using this progressive translation, their model can extract the underlying relationship between two images by continuously varying the receptive fields at different scales. All in all, TuiGAN represents a further step toward the possibility of unsupervised learning with extremely limited data. 

\begin{table*}
	\centering
	\caption{List of multi-domain I2I methods including model name, publication year, the type of training data, whether multimodal or not and corresponding insights.}
	\label{table:list_multi}
	\resizebox{\textwidth}{!}{
	\begin{tabular}{c|c|c|c|l}
		\hline
		Method & Publication & Data & Multimodal & \multicolumn{1}{c}{Insights} \\\hline
		Domain-Bank/ModularGAN & 2018 & unpaired & No & each domain each module; \\
		StarGAN & 2018 & unpaired & No & unified single model; auxiliary classifier; \\
		AttGAN & 2019 & unpaired & No & unified single model; auxiliary classifier; AE-GAN; \\
		RelGAN/STGAN & 2019 & unpaired & No & auxiliary classifier; relative-attribute; \\
		CollaGAN & 2019 & unpaired & No & auxiliary classifier; multiple inputs; \\
		\cite{lin2019image} & 2019 & unpaired & No & auxiliary domain; multi-path consistency; \\
		SGN & 2019 & unpaired & No & sym-parameter; \\
		Fixed-Point GAN & 2019 & unpaired & No & forzen discriminator; \\
		\cite{he2019deliberation} & 2019 & unpaired & No & deliberation learning; \\
		ADSPM & 2019 & unpaired & No & spontaneous motion; \\
		INIT\cite{cao2020informative} & 2020 & unpaired & No & informative sample mining network; multihop training; \\
		GANimation & 2018 & unpaired & Yes & action unit; \\
		DosGAN & 2019 & unpaired & Yes & domain classifier; \\
		UFDN & 2018 & unpaired & Yes & disentanglement; \\
		DMIT & 2019 & unpaired & Yes & disentanglement; multi-mapping; \\
		StarGANv2 & 2020 & unpaired & Yes & disentanglement; multi-task discriminator; diverseity regularization; \\
		DRIT++ & 2020 & unpaired & Yes & disentanglement; latent regression loss; \\
		GMM-UNIT & 2020 & unpaired & Yes & disentanglement; Gaussian mixture model; \\
		FUNIT & 2019 & unpaired & Yes & few-shot UI2I; multi-task discriminator;; \\
		COCO-FUNIT & 2020 & unpaired & Yes & few-shot UI2I; multi-task discriminator; content leak;\\\hline
		AGUIT & 2019 & paired+unpaired & Yes & disentanglement; domain classifier; cyclic loss; \\
		SEMIT & 2020 & paired+unpaired & No & few-shot I2I; pseudo-label; \\\hline
	\end{tabular}}
\end{table*}

\section{multi-domain Image-to-Image Translation}
\label{sec:multi-domain}
In this section, we will discuss the I2I problem on multiple domains and list correlative algorithms in Table~\ref{table:list_multi} convering model name, publication year, the type of training data, whether multi-modal or not and corresponding insights. We have discussed a series of attractive I2I works for translating two domains. However, these methods can only capture the relationship of two domains based on one model at a time. Given $n$ domains, the network requires $n\times(n-1)$ generators to be trained, which leads to an unavoidable burden. Moreover, it fails to fully use the entire training data from all the domains. Even if there exists global information learned from all the domains that can be applied to promote the translation performance, the network still only learns from two domains, and it is difficult to acquire that global multi-domain information. How to further reduce the network complexity and improve the efficiency to handle multiple domains remains unaddressed. 

Therefore, researchers study the multi-domain I2I problem. It focuses on handling multiple domains using a single unified model in which multiple outputs contain different semantic contents or style textures. We divide multi-domain I2I research into three categories: unsupervised multi-domain I2I, semi-supervised multi-domain I2I and few-shot multi-domain I2I.

\subsection{Unsupervised multi-domain Image-to-Image Translation}
In this subsection, we introduce unsupervised multi-domain I2I (multi-domain UI2I) in two aspects: single-modal output and multimodal outputs. 
\begin{figure}[!t]
\centering
\includegraphics[width=0.45\textwidth]{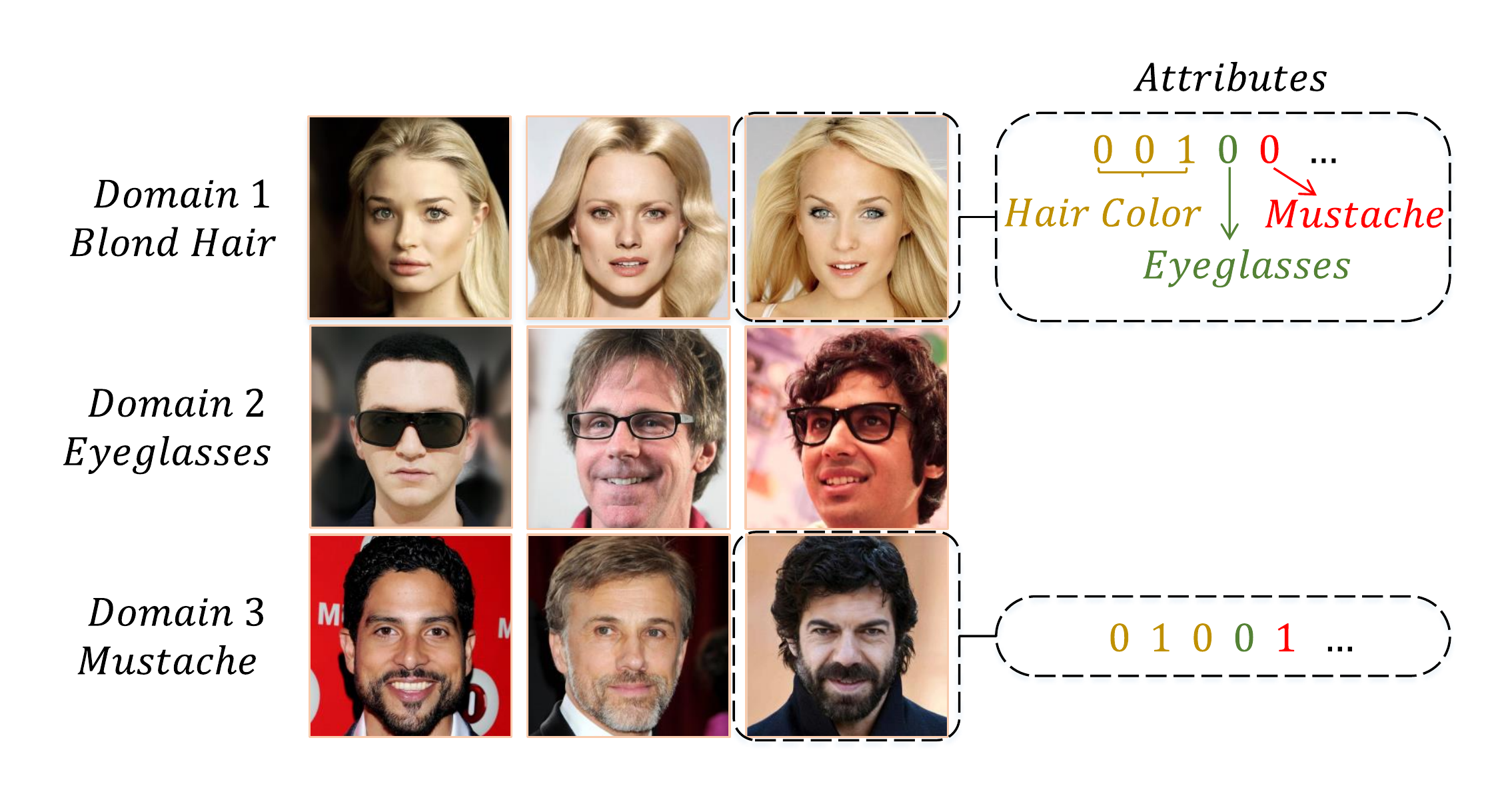}
\caption{Illustration of the dataset used in multi-domain I2I scenarios: Each domain usually means a set of images sharing the same attribute value. Images are from the CelebA dataset \cite{Liu_2015_ICCV}.}
\label{fig:cebeba}
\end{figure}
First, we explain how to achieve this translation with multiple domains. For example, the CelebA dataset \cite{Liu_2015_ICCV} contains 40 facial attributes, and each domain usually means a set of images sharing the same attribute value in \cite{Choi_2018_CVPR,8718508,Wu_2019_ICCV_relGAN,Liu_2019_CVPR}. We therefore can obtain numerous unpaired translation domains based on different attribute values, as shown in Fig.\ref{fig:cebeba}. Notwithstanding the demonstrated success of unsupervised two-domain I2I (two-domain UI2I) in subsection \ref{subsec:UI2I}, when we have multiple unpaired domains, do these two-domain UI2I methods still work? The answer may be “no.” The typical problems are the efficiency and network burden in which these two-domain UI2I can only transfer one pair of different domains through one training. The multi-domain UI2I hence attracts much attention, and we will provide detailed illustrations from two aspects: multi-domain UI2I with single-modal output and with multimodal outputs.

\subsubsection{Single-modal Output}
In addition, a large variety of multi-domain UI2I methods with a single-modal output have been proposed to obtain image representations in an unsupervised way. We classify them into three categories: training with multimodules, training with one generator and discriminator pair and training by combining knowledge in other fields.
\begin{itemize}
\item \textbf{Training with multimodules} In earlier times, methods mainly designed complex multiple modules to address multi-domain UI2I by regarding it as a composition of several two-domain UI2Is, in which each module represents each domain information. Compared with directly applied two-domain UI2I methods, these works can train all the domains at one time, which saves much training time and many model parameters.
\item \textbf{Training with one generator and discriminator pair} Unfortunately, methods training with multimodules can train translations between multiple domains at one time, but they still need to define multiple-domain modules to represent the corresponding domains. Is there a model that can train all domains at one time using the same module to process multiple-domain information? A more effective solution is to use an auxiliary label (binary or one-shot attribute vector) to represent domain information that leads to a more flexible translation. After randomly choosing the target domain label as conditional input, we can translate the source domain input to this target domain without extra translators using one generator and discriminator pair. 
\item \textbf{Training by combining knowledge in other fields} Some algorithms try to introduce knowledge from other research areas to facilitate multi-domain UI2I. To some extent, these methods have indeed brought us new insight.
\end{itemize}

\paragraph{Training with Multimodules} 
Based on the shared-latent space assumption \cite{liu2017unsupervised}, Hui et al. \cite{8545169} propose a unified framework named Domain-Bank. Given $n$ domains, Domain-Bank obtains $n$ pairs of translated results by training the network only once, while the two-domain I2I methods require the training of $n$ models for translations between different pairs of domains.
By leveraging several reusable and composable modules, Zhao et al. \cite{Zhao_2018_ECCV} propose ModularGAN to translate an image to multiple domains efficiently. They predefine an attribute set $\mathbf{A}=\{A_{1},\dots,A_{n}\}$ in which each attribute $A_{i}$ represents meaningful inherent property of each domain with different attribute values.

\paragraph{Training with One Generator and Discriminator Pair}
\begin{figure}[!t]
\centering
\includegraphics[width=0.150\textwidth]{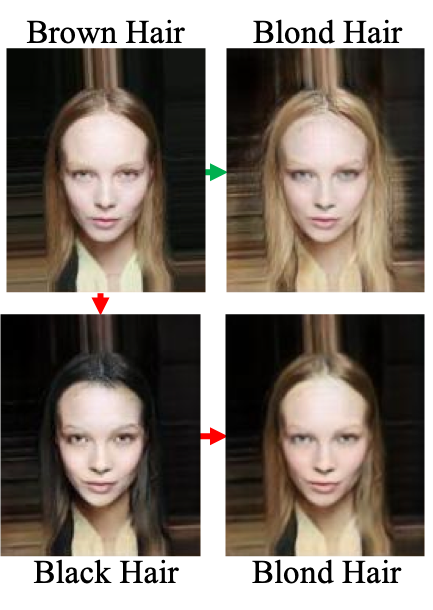}
\caption{The illustration of multipath consistency regularization \cite{lin2019image} on the translation between different hair colors. Ideally, they consider that the direct translation (i.e., one-hop translation) from brown hair to blonde should be identical to the indirect translation (i.e., two-hop translation) from brown to black to blonde.}
\label{fig:multi-path}
\end{figure}
Choi et al. propose StarGAN \cite{Choi_2018_CVPR}, which fully proves the effectiveness of the auxiliary domain label by mappings between all available domains using only a single model. They design a special discriminator and introduce an auxiliary classifier on top of it, in which the discriminator not only justifies whether an image is a natural or fake one $D_{src}(x_{A})$ but also distinguishes which domain the input belongs to $D_{cls}(x_{A})$:
\begin{equation}
D:x_{A}\mapsto{\{D_{src}(x_{A}),D_{cls}(x_{A})\}}.
\end{equation}
To translate input image $x_{A}\in A$ to target domain B, StarGAN learns an adversarial loss and a cycle-consistency loss conditioned with input domain label $c_{A}$ and target domain label $c_{B}$.

Through the three loss functions, StarGAN can achieve a scalable translation for multiple domains and obtain results with higher visual quality.
Sharing an extremely similar idea, He et al. \cite{8718508} propose AttGAN to address this problem. However, AttGAN uses an encoder-decoder architecture to model the relationship between the latent representation and the attributes.

The target-attribute constraint used in StarGAN and AttGAN fails to provide a more fine-grained control and often requires specifying the complete target attributes, even if most of the attributes are not changed. Wu et al. \cite{Wu_2019_ICCV_relGAN} and Liu et al. \cite{Liu_2019_CVPR} therefore consider a novel attribute description termed relative-attribute that represents the desired change of attributes. They propose RelGAN and STGAN, respectively, to satisfy arbitrary image attribute editing with relative attributes. Given input domain attribute $c_{A}$ and target domain attribute $c_{B}$, the relative attribute $c$ is formulated as:
\begin{equation}
c = c_{B}-c_{A}.
\end{equation}

StarGAN or AttGAN may perform worse when multiple inputs are required to obtain a desired output. Any missing input data will introduce a large bias and lead to terrible results. Therefore, CollaGAN \cite{Lee_2019_CVPR} has been proposed to process multiple-inputs from multiple domains instead of only handling single-input and single-output. 

Rather than introducing an auxiliary domain classifier, Lin et al. \cite{lin2019image} propose introducing an additional auxiliary domain and constructing a multipath consistency loss for multi-domain I2I. Their work is motivated by an important property shown in Fig. \ref{fig:multi-path}, namely, the direct translation (i.e., one-hop translation) from brown hair to blonde should ideally be identical to the indirect translation (i.e., two-hop translation) from brown to black to blonde. Their multipath consistency loss evaluates the differences between direct two-domain translation $A \mapsto{C}$ and indirect multiple-domain translations $A\mapsto{B\mapsto{C}}$ with domain $B$ as an auxiliary domain. The method regularizes the training of each task and obtains a better performance.

\paragraph{Training by Combining Knowledge in Other Fields}
By expanding the concept of a multi-domain from data to the loss area, Chang et al. \cite{Chang_2019_ICCV} introduce the sym-parameter to synchronize various mixed losses with input conditions.  
Siddiquee et al. \cite{Siddiquee_2019_ICCV} propose Fixed-Point GAN, which uses a trainable generator and a frozen discriminator to perform fixed-point translation learning. 
He et al. \cite{he2019deliberation} propose deliberation learning for I2I by adding a polishing step on the output image. 
Cao et al. \cite{cao2020informative} propose the informative sample mining network (INIT) to analyze the importance of sample selection and select the informative samples for multihop training. 
Wu et al. \cite{Wu_2019_ICCV} propose ADSPM to learn attribute-driven deformation by a spontaneous motion (SPM) estimation module and a refinement part (R) with much consideration for geometric transform.

\begin{figure}[!t]
\centering
\includegraphics[width=0.45\textwidth]{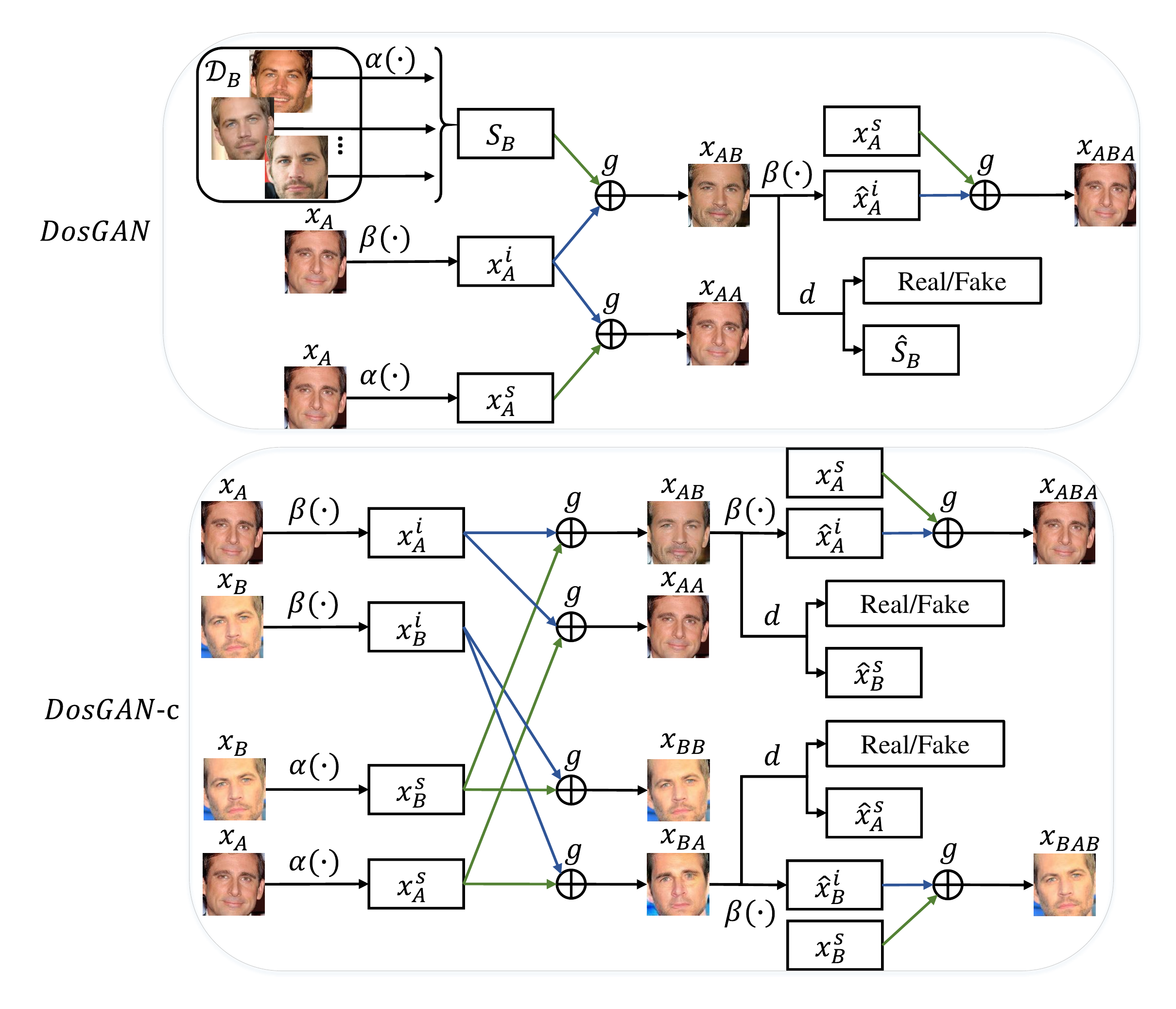}
\caption{The training architecture of DosGAN \cite{8886528}: (1) top: DosGAN for unpaired I2I; (2) bottom: DosGAN-c for unpaired conditional I2I.}
\label{fig:dosgan}
\end{figure}
\subsubsection{Multimodal Outputs}
However, all of the multi-domain approaches mentioned above still learn a deterministic mapping between two arbitrary domains. Researchers therefore consider addressing the multi-domain I2I, as well as the outputs of multimodal results.

Lin et al. observe that if the pretrained CNN network can accurately classify the domain of an image, then the output of the second-to-last layer of the classifier should well capture the domain information of this image. Combining this network with a domain classifier, they propose domain-supervised GAN (DosGAN) \cite{8886528} to use the domain label as an explicit supervision and pretrain a deep CNN to predict which domain an image is from. 
The detailed training architecture is shown in Fig.\ref{fig:dosgan}.

In addition, GANimation \cite{Pumarola_2018_ECCV} is proposed to generate anatomically aware facial animation. It continuously synthesizes anatomical facial movements by controlling the magnitude of activation of each action unit (AU).

By exploiting the disentanglement assumption, UFDN \cite{liu2018unified}, DMIT\cite{yu2019multi}, StarGAN v2 \cite{choi2020stargan}, DRIT++ \cite{lee2018diverse} and GMM-UNIT \cite{liu2020gmm} are proposed to perform multimodal outputs in a multi-domain UI2I setting. For example, DRIT++ consists of two content encoders $\{E_{A}^{c},E_{B}^{c}\}$, two attribute encoders $\{E_{A}^{a},E_{B}^{a}\}$, two generators $\{G_{A},G_{B}\}$, two discriminators $D_{A},D_{B}$ and a content discriminator $D_{adv}^{c}$. Through weight sharing and a content discriminator $D_{adv}^{c}$, the network can achieve representation disentanglement with content adversarial loss. Then, it leverages cross-cycle consistency loss for forward and backward translations. In addition to these two losses, these methods also use domain adversarial loss, self-reconstruction loss, latent regression loss and an extra mode-seeking regularization to effectively improve the sample diversity and visual quality. 

\subsection{Semi-Supervised multi-domain Image-to-Image Translation}
Li et al. \cite{li2019attribute} propose an attribute guided I2I (AGUIT) model that is the first work to handle multimodal and multi-domain I2I with semi-supervised learning. AGUIT is trained following three steps. The first step is representation decomposition, which extracts content and style features with two encoders, a content discriminator and a label predictor, and the style code includes a noise part and an attribute part. The second step is reconstruction and translation using AdaIN \cite{huang2017arbitrary} and a discriminator and a domain classifier. The third step is consistency reconstruction with cycle consistency loss and feature consistency loss. AGUIT is trained in a training set containing labeled images mixed with unlabeled images so that it can translate attributes well. 
By going one step further to reduce the amount of labeled data required in the training process and source domain, Wang et al. \cite{Wang_2020_CVPR} propose SEMIT to address the challenging problem combined with few-shot I2I. SEMIT initially applies semi-supervised learning via a noise-tolerant pseudo-labeling procedure to assign pseudo-labels to the unlabeled training data. Then, it performs UI2I using adversarial loss, classification loss and reconstruction loss with only a few labeled examples during training.

\subsection{Few-Shot multi-domain Image-to-Image Translation}

\begin{figure}[!t]
	\centering
	\includegraphics[width=0.45\textwidth]{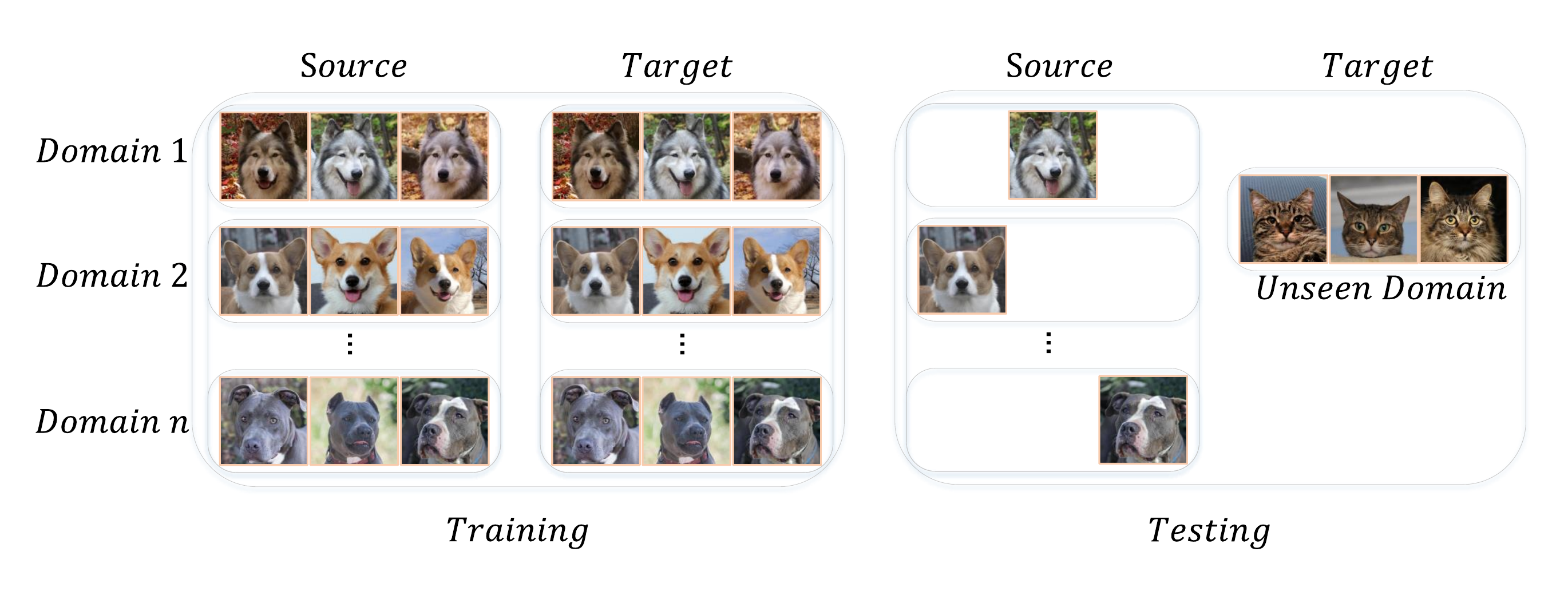}
	\caption{Illustration of the dataset used in few-shot multi-domain I2I scenarios: The training set consists of multiple domains in which the source and target images are randomly sampled from arbitrary $n$ domains; given very few images of the target unseen domain (unseen in the training process), few-shot multi-domain I2I aims to translate a source content image (randomly sampled from $n$ domains) into an image analogous to this unseen domain.}
	\label{fig:few-shot-multi-domain}
\end{figure}

Although prolific, the aforementioned successful multi-domain I2I techniques can hardly rapidly generalize from a few examples. 
In contrast, humans can learn new tasks rapidly using what they learned in the past. Given a static picture of a butterfly, you can easily imagine it flying similar to a bird or a bee after watching a video of a flock of birds or a swarm of bees in flight. Hence, few-shot multi-domain I2I attracts much attention.

Liu et al. \cite{Liu_2019_ICCV} seek a few-shot UI2I algorithm, FUNIT, to successfully translate source images to analogous images of the target class with many source class images but few target class images available. FUNIT first trains a multiclass UI2I with multiple classes of images, such as those of various animal species, based on a few-shot image translator and a multitask adversarial discriminator. In the test time, it can translate any source class image to analogous images of the target class with a few images from a novel object class (namely, the unseen target class).

However, FUNIT fails to preserve domain invariant appearance information in the content image because of the severe influence of the style code extracted from the target image, namely, the \textit{content loss} problem. Saito et al. \cite{saito2020coco} therefore proposed COCO-FUNIT to redesign a content-conditioned style encoder that interpolates content information into a style code.

\begin{figure}[!t]
\centering
\includegraphics[width=0.45\textwidth]{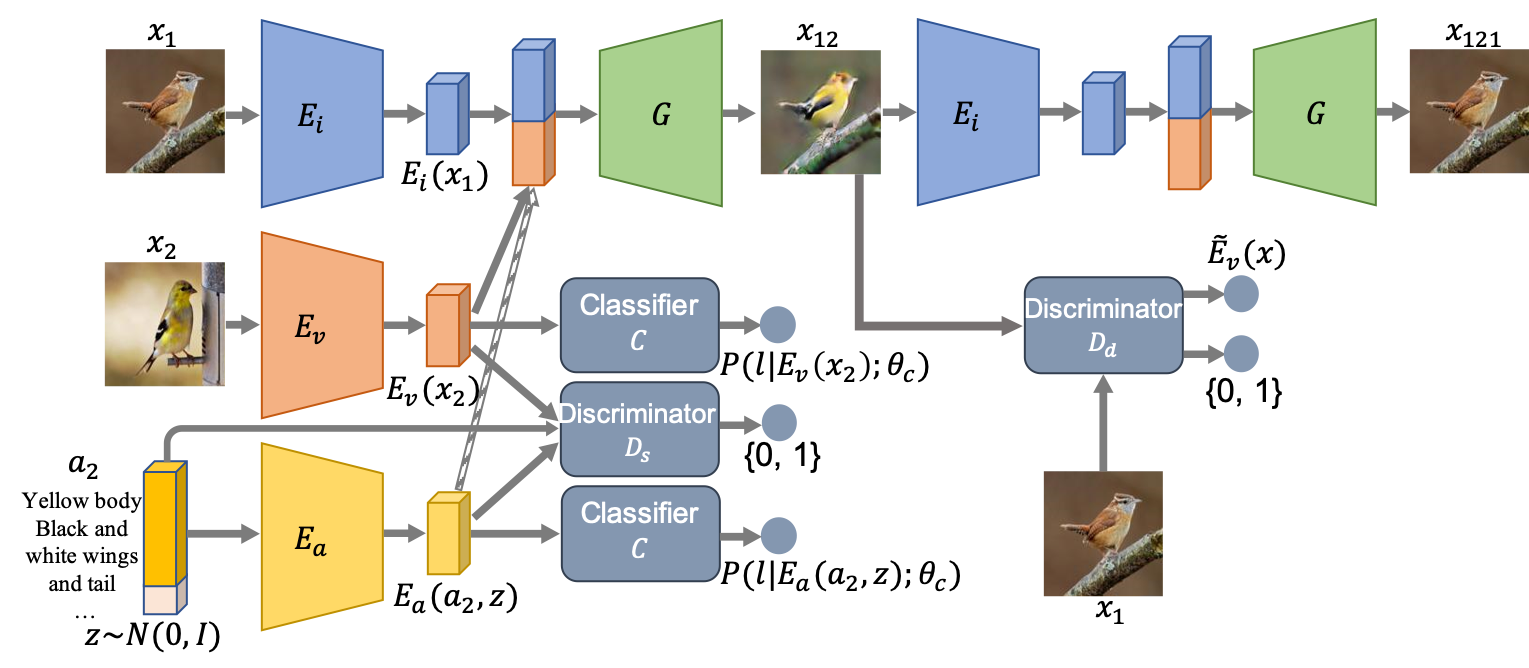}
\caption{The architecture of ZstGAN \cite{lin2019zstgan}.}
\label{fig:zstgan}
\end{figure}
Moreover, Lin et al. found that the current I2I works translate from random noise, which, unlike humans, cannot easily adapt acquired prior knowledge to solve new problems. They hence proposed the unsupervised zero-shot I2I (ZstGAN) \cite{lin2019zstgan} shown in Fig.\ref{fig:zstgan}. ZstGAN uses meta-learning to transfer translation knowledge from seen domains to unseen classes using a translator trained on seen domains to translate images of unseen domains with annotated attributes.

\begin{figure}[!t]
	\centering
	\includegraphics[width=0.45\textwidth]{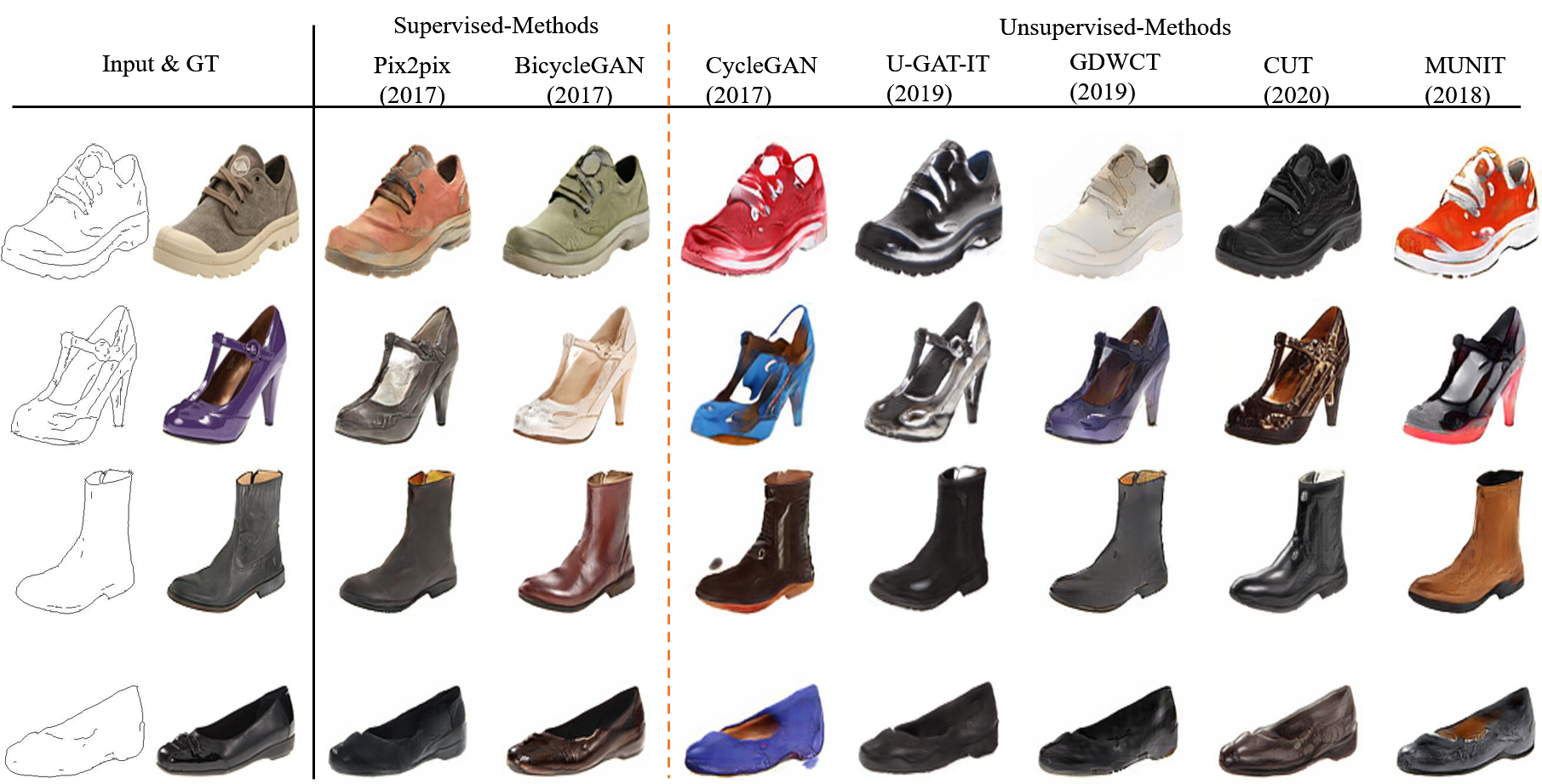}
	\caption{Qualitative comparison on single modal two-domain I2I methods. Here we show the examples of edge$\to$shoes.}
	\label{fig:singlemodal2domain}
\end{figure}

\begin{figure}[!t]
	\centering
	\includegraphics[width=0.45\textwidth]{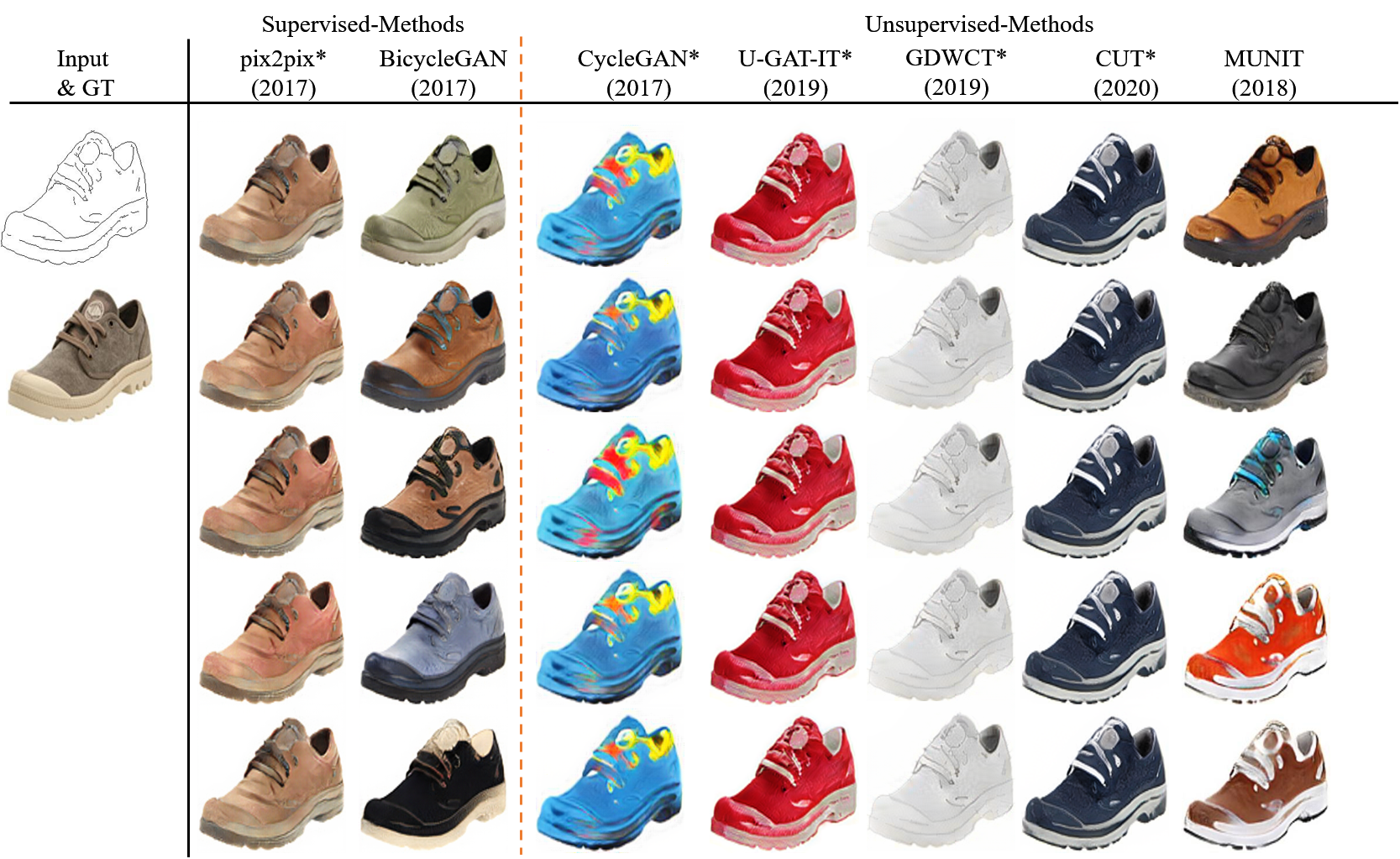}
	\caption{Qualitative comparison on multi-modal two-domain I2I methods. Here we show the examples of edge$\to$shoes, where * indicates addtionally injecting noise vectors to the translation network.}
	\label{fig:multimodal2domain}
\end{figure}

\section{Experimental evaluation}
In this section, we evaluate twelve I2I models on two tasks, including seven two-domain algorithms on edge-to-shoes translation task and five multi-domain algorithms on attribute manipulation task. We train all the models following their default settings as original papers except the same dataset and implementation environments. The selection criteria of methods mainly takes into account algorithm categories and publication years. All experimental codes come from the public official version.
\subsection{Datasets}
\noindent{\textbf{UT-Zap50K}}
We utilize the UT-Zap50K dataset~\cite{yu2014fine} to evaluate the performance of two-domain I2I methods. The number of training pairs is 49826 where each pair consists of a shoes image and its corresponding edge map. And the number of testing images is 200. We resize all images to 256$\times$256 for training and testing. In unsupervised setting, images from source domain and target domain are not paired.

\noindent{\textbf{CelebA}}
We employ the CelebFaces Attributes (CelebA) dataset~\cite{Liu_2015_ICCV} to compare the performance of multi-domain I2I methods. It contains 202,599 face images of celebrities with 40 \textit{with/without} attribute labels for each image. We randomly divide all images into training set, validation set and test set with ratio $8:1:1$. Next, we center-crop the initial 178$\times$218 size images to 178$\times$178. Finally, after resizing all images to 128$\times$128 by bicubic interpolation, we construct the multiple domains dataset using the following attributes: Black hair, Blond hair, Brown hair, gender (male/female), and age (young/old).
\subsection{Metrics}
We evaluate both the visual quality and the diversity of generated images using Frechét inception distance (FID), Inception score (IS) and Learned Perceptual Image Patch Similarity (LPIPS). 

\noindent{\textbf{Fr\'{e}chet inception distance (FID)} \cite{heusel2017gans}} is computed by measuring the mean and variance distance of the generated and real images in a deep feature space. A lower score means a better performance. (1) For single-modal two-domain setting, we directly compared the mean and variance of generated and real sets. (2) For multi-modal two-domain setting, we sample the same testing set 19 times. Then compute the FID for each testing set and average the scores to get the final FID score. (3) For single-modal multi-domain setting, we compute the FID score in each domain and then average the scores. (4) For multi-modal multi-domain setting, we first sample each image in each domain 19 times. Then we compute the average FID scores within each domain. Finally, we average the scores again to get the result.

\noindent{\textbf{Inception score (IS)}}~\cite{salimans2016improved} encodes the diversity across all translated outputs. It exploits a pretrained inception classification model to predict the domain label of the translated images. A higher score indicates a better translated performance. The evaluation process is just as similar as FID.

\noindent{\textbf{Learned Perceptual Image Patch Similarity (LPIPS)}}~\cite{zhang2018unreasonable} evaluates the diversity of the translated images and is demonstrated to correlate well with human perceptual similarity. It is computed as the average LPIPS distance between pairs of randomly sampled translation outputs from the same input. Specifically, we sample 100 images with 19 pairs of outputs (randomly sample two style vectors or inputs added random Guassian noise). We then compute the distance between two generated results and get an average. A higher LPIPS score means a more realistic, diverse translated result.

\begin{figure}[!t]
	\centering
	\includegraphics[width=0.45\textwidth]{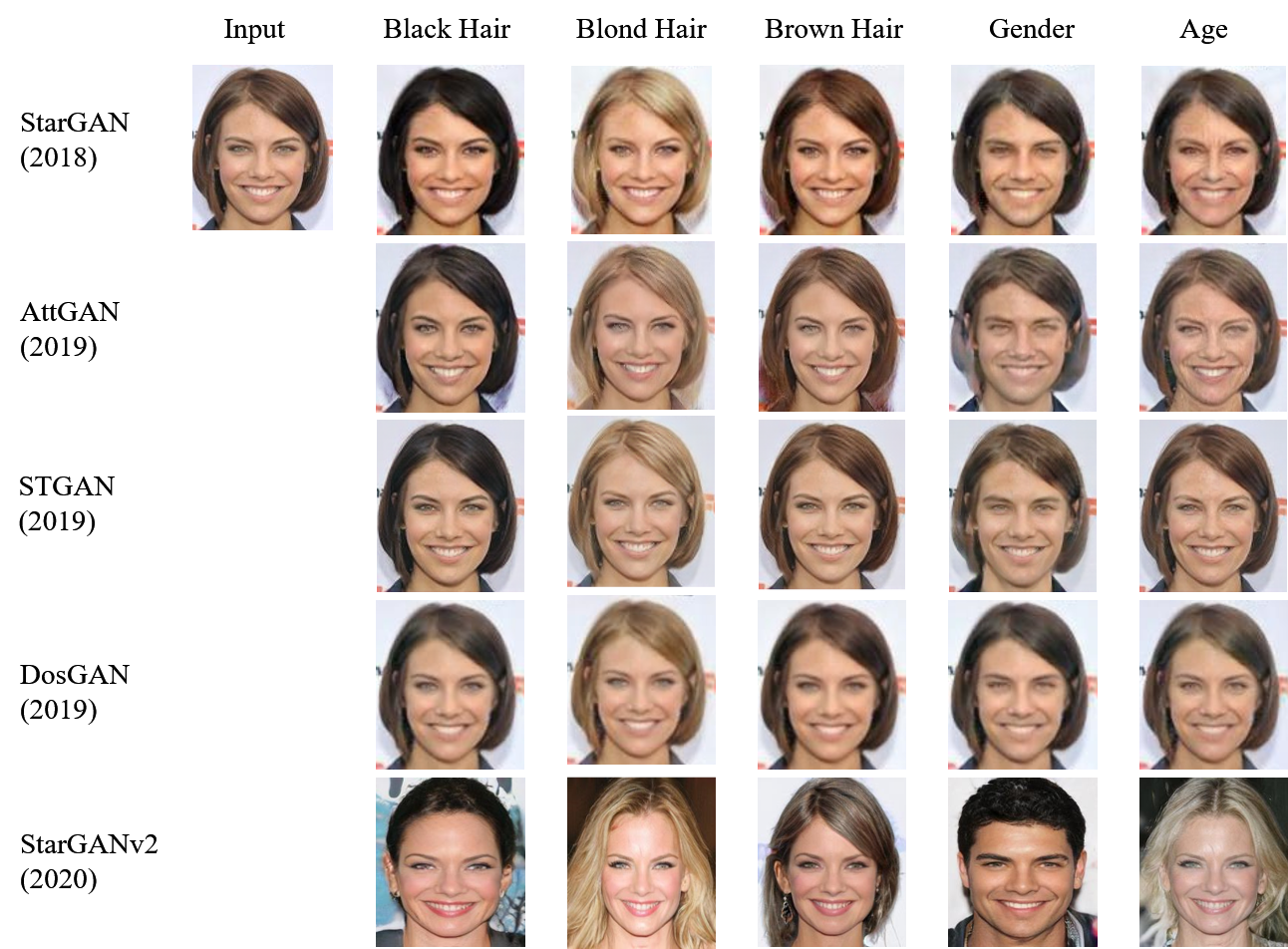}
	\caption{Qualitative comparison on single modal multi-domain I2I methods. Here we show the examples of 5 attributes.}
	\label{fig:singlemodalmultidomain}
\end{figure}

\subsection{Results} 
A fair comparison is only possible by keeping all the parameters consistent. That said, it is difficult to declare that one algorithm has an absolute superiority over the others. Besides model design itself, there are still many factors influencing the performance, such as training time, batch size and iteration times, FLOPs and number of parameters, etc. Therefore, our conclusion only build on current experimental settings, models and tasks. 


\begin{figure}[!t]
	\centering
	\includegraphics[width=0.45\textwidth]{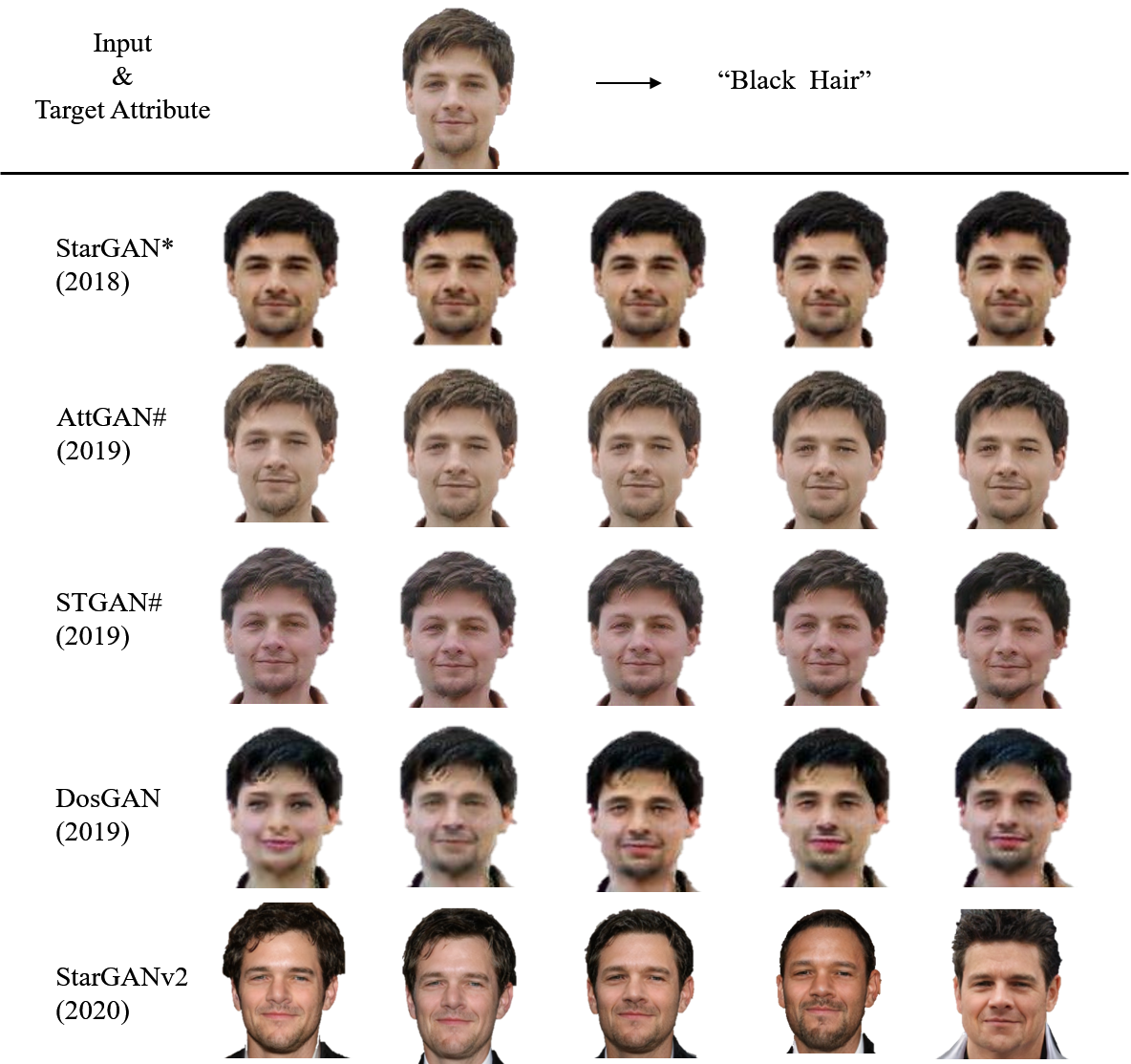}
	\caption{Qualitative comparison on multi-modal two-domain I2I methods. Here we show the examples of 5 domain translation, where * indicates additionally injecting noise vectors to the translation network, $\sharp$ denotes the linear interpolation between two attributes:  Brown-Hair$\to$Black-Hair.}
	\label{fig:multimodalmultidomain}
\end{figure}

\noindent{\textbf{Two-domain I2I}} We qualitatively and quantitatively compare pix2pix~\cite{isola2017image}, BicycleGAN~\cite{zhu2017toward}, CycleGAN~\cite{zhu2017unpaired}, U-GAT-IT~\cite{kim2019u}, GDWCT~\cite{cho2019image}, CUT~\cite{park2020contrastive} and MUNIT~\cite{huang2018multimodal} in single-modal and multi-modal setting respectively. 

The single-modal qualitative comparisons are shown in Fig.~\ref{fig:singlemodal2domain} where two supervised methods pix2pix and BicycleGAN achieve better FID, IS and LPIPS scores than unsupervised methods CycleGAN, U-GAT-IT and GDWCT. Without any supervision, the newest method CUT gets the best FID and IS scores in Table.~\ref{table:two-domain} than the rest methods including supervised and unsupervised. There could be a couple reasons for that. First, the backbone of CUT, namely StyleGAN, is a strong powerful GAN model for image synthesis compared to others. Besides, the contrastive learning they used is an effective content constraint for translation. 

As for the multi-modal setting shown in Fig.~\ref{fig:multimodal2domain}, we inject the Gaussian noise into the input of pix2pix, CycleGAN, U-GAT-IT, GDWCT and CUT to get multi-modal results. However, they can hardly generate the diverse outputs. On the contrary, the multi-modal algorithms BicycleGAN and MUNIT can acquire multi-modal and realistic results. Also supervised method BicycleGAN achieves 0.047 more LPIPS scores than unsupervised method MUNIT as shown in Table.~\ref{table:two-domain}.

\begin{table*}
	\centering
	\caption{The average FID, IS, LPIPS scores of different two-domain I2I methods trained on UT-Zap50K dataset~\cite{yu2014fine} in task edge$\to$shoes. The best scores are in bold.}
	\label{table:two-domain}
	\resizebox{0.90\textwidth}{!}{
		\begin{tabular}{c|c|c|c|c|c|c|c}
			\multirow{2}{*}{Category} & \multicolumn{2}{|c|}{Supervised I2I} & \multicolumn{5}{|c}{Unsupervised I2I}       \\\cline{2-8}
			& single-modal     & multi-modal     & \multicolumn{4}{|c|}{single-mdoal} & multi-modal \\ \hline
			Method                    & pix2pix          & BicycleGAN      & CycleGAN & U-GAI-IT & GDWCT & CUT & MUNIT       \\\hline
			Publication               & 2017             & 2017            & 2017     & 2019     & 2019 & 2020& 2018         \\\hline
			FID ($\downarrow$)       &65.09            & 64.23           & 73.76    & 91.33    & 79.56& \textbf{50.03}& 75.31  \\\hline
			IS ($\uparrow$)        &3.08$\pm$0.39    & 2.96$\pm$0.36    & 2.66$\pm$0.25  & 2.86$\pm$0.31 & 2.69$\pm$0.39  &\textbf{3.21$\pm$0.77}     & 2.33$\pm$0.25     \\\hline
			LPIPS ($\uparrow$) & 0.064 &  \textbf{0.237}  &  0.038  &  0.028  & 0.017  & 0.019  & 0.190         \\\hline
	\end{tabular}}
\end{table*}

\begin{table*}
	\centering
	\caption{The average FID, IS, LPIPS scores of different multi-domain I2I methods trained on CelebA dataset~\cite{Liu_2015_ICCV} in 5 domains including Black hair, Blond hair, Brown hair, gender (male or female) and age (young or old). In addition to the final average metric scores, we also report two domains results (Black hair and gender) for reference. The best scores are in bold.}
	\label{table:multi-domain}
	\resizebox{\textwidth}{!}{
		\begin{tabular}{c|c|c|c|c|c|c|c|c|c|c|c|c}
			\multicolumn{2}{c}{Category}                                                                                                                            & Method    & Publication & \begin{tabular}[c]{@{}c@{}}FID\\ (Black Hair)\end{tabular} & \begin{tabular}[c]{@{}c@{}}FID\\ (Gender)\end{tabular} & \begin{tabular}[c]{@{}c@{}}FID\\($\downarrow$)\end{tabular} & \begin{tabular}[c]{@{}c@{}}IS\\ (Black Hair)\end{tabular} & \begin{tabular}[c]{@{}c@{}}IS\\ (Gender)\end{tabular} & \begin{tabular}[c]{@{}c@{}}IS\\ ($\uparrow$)\end{tabular} & \begin{tabular}[c]{@{}c@{}}LPIPS\\ (Black Hair)\end{tabular} & \begin{tabular}[c]{@{}c@{}}LPIPS\\ (Gender)\end{tabular} & \begin{tabular}[c]{@{}c@{}}LPIPS\\ ($\uparrow$)\end{tabular} \\\hline
			\multirow{5}{*}{\begin{tabular}[c]{@{}c@{}}Unsupervised \\ I2I\end{tabular}} & \multirow{3}{*}{\begin{tabular}[c]{@{}c@{}}single-\\ modal\end{tabular}} & StarGAN   & 2018        & 101.66   & 96.96  & 94.39  & 1.494$\pm$0.167  & 1.506$\pm$0.295  & 1.497$\pm$0.158   & 0.009 & 0.010   & 0.011   \\\cline{3-13}
			&   & AttGAN    & 2019  & 87.80  & 83.29    & 74.48   & 1.136$\pm$0.056   & 1.228$\pm$0.086  & 1.231$\pm$0.121  & 0.023    & 0.027   & 0.021 \\\cline{3-13}
			&   & STGAN     & 2019  & 86.80  & 95.37    & 82.41  & 1.417$\pm$0.205  & 1.636$\pm$0.461 & 1.568$\pm$0.304 & 0.065   & 0.043 & 0.036  \\\cline{2-13}
			& \multirow{2}{*}{\begin{tabular}[c]{@{}c@{}}multi-\\ modal\end{tabular}} & DosGAN &  2019 & 68.38  & 71.35 & 73.36   & 1.617$\pm$0.285  & 1.556$\pm$0.284  & 1.568$\pm$0.206 & 0.064 & 0.055 & 0.061  \\\cline{3-13}
			 & & StarGANv2 &  2020        & 42.89  & 43.86   & \textbf{40.52}   & 1.537$\pm$0.233  & 1.674$\pm$0.440   & \textbf{1.586$\pm$0.278}     & 0.414 & 0.377  & \textbf{0.397} \\\hline                                                    
	\end{tabular}}
\end{table*}

\noindent{\textbf{Multi-domain I2I}} We qualitatively and quantitatively compare StarGAN~\cite{Choi_2018_CVPR}, AttGAN~\cite{8718508}, STGAN~\cite{Liu_2019_CVPR} DosGAN~\cite{8886528} and StarGANv2~\cite{choi2020stargan} in single-modal and multi-modal setting respectively. 

In Fig.~\ref{fig:singlemodalmultidomain}, all methods can successfully achieve multiple domains translation. However, StarGAN and AttGAN generate obvious visible artifacts while DosGAN leads to blurry results. The results of STGAN are pretty excellent whereas StarGANv2 can generate realistic and vivid translated results by changing the image style latent code. Table~\ref{table:multi-domain} shows that StarGANv2 acquires the best FID and IS scores. Similarly, there are also many factors contributed to such result including stronger GAN backbone, more effective training strategies, higher quality dataset, etc.

We also conduct the multimodal multi-domain I2I experiments for comparison. In detail, we additionally inject noise vectors to StarGAN for multi-modal translation. As for AttGAN and STGAN, we apply the linear interpolation between two attributes: Brown-Hair$\to$Black-Hair as multi-modal results. As shown in Fig~\ref{fig:multimodalmultidomain}
 and Table~\ref{table:multi-domain}, StarGAN fails to generate diverse outputs and get the worst LPIPS score despite the injected randomness. AttGAN and STGAN can generate multi-modal results, but the gap is very small meanwhile DosGAN performs worse translation quality. In comparison, StarGANv2 can generate totally different modality leading to the best LPIPS score.

\noindent{\textbf{Conclusion}} 
Generally, supervised methods usually produce better translated results than unsupervised methods on similar network structure. However, in some special cases, supervised methods do not always perform better than unsupervised methods such as CUT which benefits from the development of network architecture (StyleGAN) and more effective training strategies (contrastive learning). As reported in Table.~\ref{table:two-domain} and Table.~\ref{table:multi-domain}, choosing an updated model to train I2I task may be a good idea because such model is usually trained with some of the latest training strategies and well-designed network architecture. Moreover, the high-quality dataset plays a crucial role in I2I task.

\section{Application}
\label{sec:application}

In this section, we review the various and fruitful applications of I2I shown in Table \ref{table:UI2I}. We classify the main applications following the taxonomy of I2I methods.

\begin{table*}
	\centering
	\caption{Applications of I2I discussed in Section \ref{sec:application}}
	\label{table:UI2I}
	\resizebox{\textwidth}{!}{
	\begin{tabular}{c|c|c|c|c|c|c|c|c|c|c|c}
		\multirow{3}{*}{TASK} & \multicolumn{6}{|c|}{TWO-DOMAIN I2I} & \multicolumn{5}{|c}{MULTI-DOMAIN I2I} \\\cline{2-12}
		& \multicolumn{2}{|c|}{Supervised I2I} & \multicolumn{2}{|c|}{Unsupervised I2I} & Semi-supervised I2I & Few-shot I2I & \multicolumn{2}{|c|}{Unsupervised I2I} & \multicolumn{2}{|c|}{Semi-supervised I2I} & Few-shot I2I \\\cline{2-12}
		& \begin{tabular}[c]{@{}l@{}}Single-modal\\ Output\end{tabular}  & \begin{tabular}[c]{@{}l@{}}Multi-modal\\ Outputs\end{tabular}
		& \begin{tabular}[c]{@{}l@{}}Single-modal\\ Output\end{tabular}  & \begin{tabular}[c]{@{}l@{}}Multi-modal\\ Outputs\end{tabular}
		& \begin{tabular}[c]{@{}l@{}}Single-modal\\ Output\end{tabular}  & \begin{tabular}[c]{@{}l@{}}Single-modal\\ Output\end{tabular}
		& \begin{tabular}[c]{@{}l@{}}Single-modal\\ Output\end{tabular}  & \begin{tabular}[c]{@{}l@{}}Multi-modal\\ Outputs\end{tabular}
		& \begin{tabular}[c]{@{}l@{}}Single-modal\\ Output\end{tabular}  &
		\begin{tabular}[c]{@{}l@{}}Multi-modal\\ Outputs\end{tabular}
		  & \begin{tabular}[c]{@{}l@{}}Single-modal\\ Output\end{tabular} \\\hline
		Semantic synthesis 
			& {\begin{tabular}[c]{@{}l@{}}
				\cite{hertzmann2001image,isola2017image},\\
				 \cite{Wang_2018_ECCV,wang2018high},\\
				 \cite{Park_2019_CVPR,Zhu_2020_CVPR,regmi2018cross}, \\ 
				 \cite{zhang2020cross,zhou2020full} \end{tabular}}   
			& \cite{zhu2017toward}  
			&{\begin{tabular}[c]{@{}l@{}} 
					\cite{Yi_2017_ICCV,zhu2017unpaired},\\
					\cite{li2018unsupervised,park2020contrastive},\\
			 		\cite{chen2020distilling,li2019asymmetric} \end{tabular}}  
			& {\begin{tabular}[c]{@{}l@{}}
					\cite{kazemi2018unsupervised,mao2019mode},\\
					\cite{alharbi2019latent}  \end{tabular}}
			&-  & \cite{lin2019learning,benaim2018one} & \cite{he2019deliberation}& - &-  & -&- \\\hline
		{\begin{tabular}[c]{@{}l@{}}Person image synthesis\\(skeleton-to-person)\\(virtual try-on) \end{tabular}} & \cite{zhang2020cross,zhou2020full} & - &
		{\begin{tabular}[c]{@{}l@{}} 
				\cite{ma2018gan,10.1145/3123266.3123277},\\
				\cite{NIPS2017_6644,siarohin2018deformable},\\
				\cite{Ma_2018_CVPR,Esser_2018_CVPR},\\
				\cite{Dong_2019_ICCV} \end{tabular}}
				&-  & - & - &- &-&-  &-  &-\\\hline
		Sketch-to-image 
			& {\begin{tabular}[c]{@{}l@{}}\cite{isola2017image,Wang_2018_ECCV},\\
					\cite{wang2018high,zhang2020cross}, \\ 
					\cite{zhou2020full} \end{tabular}} 
			& {\begin{tabular}[c]{@{}l@{}}\cite{zhu2017toward,bansal2018pixelnn},\\
					\cite{huang2020semantic} \end{tabular}}
			&{\begin{tabular}[c]{@{}l@{}} 
					\cite{Yi_2017_ICCV,chen2018sketchygan},\\
					\cite{kim2017learning,zhu2017unpaired}, \\
					\cite{amodio2019travelgan,Chen_2020_CVPR} \end{tabular}} 
			& {\begin{tabular}[c]{@{}l@{}} \cite{kazemi2018unsupervised,pmlr-v80-almahairi18a},\\
					\cite{huang2018multimodal,lee2018diverse},\\
					\cite{alharbi2019latent}\end{tabular}} 
			&\cite{li2020staged}  & -
			& \cite{cao2020informative} & \cite{liu2020gmm}& -& - &- \\\hline
		Paint-to-image & -& - & \cite{park2020swapping,tomei2019art2real}  & \cite{lee2018diverse} & - & \cite{benaim2018one} &- & -& - & - & -\\\hline
		Text-to-image & -& - & \cite{ma2018gan} & \cite{mao2019mode}& - &- & - & \cite{yu2019multi} &-  &- &\cite{lin2019zstgan} \\\hline
		{\begin{tabular}[c]{@{}c@{}}Semantic\\ manipulation \end{tabular}}
			& {\begin{tabular}[c]{@{}l@{}} \cite{Park_2019_CVPR,Zhu_2020_CVPR},\\ \cite{lee2020maskgan} \end{tabular}} & -
			& {\begin{tabular}[c]{@{}l@{}} \cite{Shocher_2020_CVPR,park2020swapping},\\ \cite{chang2018pairedcyclegan} \end{tabular}}
			& -& -& -& -& -& -& - &- \\\hline
		{\begin{tabular}[c]{@{}c@{}}Attribute \\manipulation \end{tabular}} & - & -
		& {\begin{tabular}[c]{@{}l@{}} \cite{liu2017unsupervised,amodio2019travelgan},\\ \cite{zhao2020unpaired,park2020swapping},\\
			\cite{cho2019image,emami2020spa} \end{tabular}} & \cite{ma2018exemplar} &- & \cite{cohen2019bidirectional} 
		& {\begin{tabular}[c]{@{}l@{}}\cite{8545169,Zhao_2018_ECCV},\\
					\cite{Choi_2018_CVPR,8718508},\\
					\cite{Wu_2019_ICCV_relGAN,Liu_2019_CVPR},\\
					\cite{Lee_2019_CVPR,lin2019image},\\
					\cite{Siddiquee_2019_ICCV,cao2020informative},\\
		    		\cite{Wu_2019_ICCV} \end{tabular}} 
		& {\begin{tabular}[c]{@{}l@{}}
					\cite{8886528,Pumarola_2018_ECCV},\\
					\cite{liu2018unified,yu2019multi},\\
					\cite{choi2020stargan,liu2020gmm}
			\end{tabular}} 
		&\cite{li2019attribute,Wang_2020_CVPR} & \cite{li2019attribute} & \cite{Liu_2019_ICCV,saito2020coco}\\\hline
		Retargeting & - & - & \cite{Bansal_2018_ECCV} & - & -& -& -& -& -& - &-\\\hline
		Image inpainting& -& - 
			&{\begin{tabular}[c]{@{}l@{}} 
				\cite{Pathak_2016_CVPR,zhu2017toward},\\
				\cite{Song_2018_ECCV,Liu_2019_ICCV},\\
				\cite{Zhao_2020_CVPR}\end{tabular}} & -& -& -& - &- & -& - &-\\\hline
		Image outpainting& -& - & \cite{Zhang_2020_CVPR} & -& -& -& - &- & -& - &-\\\hline
	 	Image-to-cartoon& -& - & 
	 		{\begin{tabular}[c]{@{}l@{}} 
	 			\cite{DBLP:conf/iclr/TaigmanPW17,Yi_2017_ICCV},\\
	 			\cite{kim2019u,gokaslan2018improving},\\
	 			\cite{zhao2020unpaired,ma2018gan},\\
	 			\cite{shi2019warpgan,wang2020learning},\\
	 			\cite{chen2018cartoongan,zheng2019unpaired} \\ \end{tabular}}
 			 &	\cite{lee2018diverse,chang2020domainspecific} & - & \cite{lin2020tuigan} & - & \cite{liu2018unified,lee2020drit++} & -& - &-\\\hline
	 	Image-to-comics& -& - & \cite{pkesko2019comixify,su2021mangagan} & - & -& -& - &- & -& - &-\\\hline
	 	{\begin{tabular}[c]{@{}c@{}}Chinese\\ character translation\end{tabular}}& - &-& \cite{gao2020gan}& - &- & -& -& -& -& - & - \\\hline
	 	style transfer & artistic:\cite{hertzmann2001image,isola2017image}  & artistic:\cite{zhu2017toward}  				
	 		&{\begin{tabular}[c]{@{}l@{}}
	 			artistic:\cite{zhu2017unpaired},\\
	 			\cite{kim2019u,jiang2020tsit},\\
	 			\cite{park2020contrastive,cho2019image};
	 			\\photo-realistic:\\
	 			\cite{Yi_2017_ICCV,park2020swapping},\\
	 			\cite{zhu2017unpaired,liu2017unsupervised},\\
	 			\cite{jiang2020tsit,park2020contrastive}\end{tabular}}  
 			& {\begin{tabular}[c]{@{}l@{}}
 				artistic:\cite{chang2020domainspecific};\\
 				photo-realistic:\\
 				\cite{huang2018multimodal,lee2018diverse},\\
 				\cite{ma2018exemplar,ma2018exemplar},\\
 	    		\cite{shen2019towards,chang2020domainspecific},\\
 	    		\cite{alharbi2019latent}\end{tabular}} 
 			& - 
 			& {\begin{tabular}[c]{@{}l@{}}
 					artistic:\cite{benaim2018one},\\
 					\cite{lin2020tuigan,Chang_2019_ICCV};\\
 			 		photo-realistic:\\
 			 		\cite{benaim2018one,cohen2019bidirectional},\\
 			 		\cite{lin2020tuigan}\end{tabular}} 
 			& {\begin{tabular}[c]{@{}l@{}}
 					artistic:\cite{8545169},\\
 					\cite{8718508,he2019deliberation};\\
 					photo-realistic:\\
 					\cite{8718508,Liu_2019_CVPR},\\
 					\cite{he2019deliberation}\end{tabular}} 
 			&  {\begin{tabular}[c]{@{}l@{}}
 					artistic:\cite{lee2020drit++};\\
 					photo-realistic:\\
 					\cite{8886528,yu2019multi},\\
 					\cite{lee2020drit++}
 				\end{tabular}}
 			& -& -& -\\\hline
	 	 {\begin{tabular}[c]{@{}c@{}}image\\ super-resolution\end{tabular}} & \cite{hertzmann2001image} & \cite{bansal2018pixelnn}& -& - & \cite{mustafa2020transformation}& -& -& -& -& - & -\\\hline
	 	image denoising & \cite{armanious2020medgan} &-  &
	 		{\begin{tabular}[c]{@{}l@{}}
	 			\cite{manakov2019noise},\\
	 			\cite{touvron2020powers}\end{tabular}} &- & \cite{mustafa2020transformation}& -& - &- & - &- &-\\\hline
	 	image deraining & \cite{zhang2019image,li2019heavy}
	 			& - & \cite{zhu2019singe}& -& -& -& -& -& -& - & - \\\hline
	 	image dehazing& \cite{dudhane2019ri} &  -
	 			& {\begin{tabular}[c]{@{}l@{}}
	 			\cite{engin2018cycle,cho2019dehazegan},\\
	 			\cite{chen2019image,cho2020underwater}\end{tabular}}& -& -& -& -& -& -& - & -\\\hline
	 	image deblurring & {\begin{tabular}[c]{@{}l@{}}\cite{kupyn2018deblurgan,Liu_2018_ECCV_Workshops},\\
	 			\cite{kupyn2019deblurgan}\end{tabular}} & -
	 			&\cite{madam2018unsupervised}
	 			& -& -& -& -& -& -& - & - \\\hline
	 	image colorization&{\begin{tabular}[c]{@{}l@{}} 
	 			\cite{isola2017image,Wang_2018_ECCV},\\
	 			\cite{zhang2017real,he2018deep},\\
	 			\cite{zhang2019deep,xu2020stylization} \end{tabular}}& - &{\begin{tabular}[c]{@{}l@{}} \cite{ma2018gan,suarez2017infrared},\\
 			\cite{Lee_2020_CVPR} \end{tabular}}& - & \cite{mustafa2020transformation}& -& -& -& -& - & - \\\hline
	 	{\begin{tabular}[c]{@{}c@{}}image quality\\ improvement\end{tabular}}&{\begin{tabular}[c]{@{}c@{}} \cite{ignatov2017dslr},\\
	 	\cite{de2018fast}\end{tabular}}& - & \cite{zhu2017unpaired,chen2018deep}
	 	& -& -& -& -& -& -& - & -\\\hline
	 	{\begin{tabular}[c]{@{}c@{}}wildlife\\ habitat analysis\end{tabular}} &\} & - & \cite{zheng2019exploiting} & -& -& -& -& -& -& - & -\\\hline
	 	disease diagnosis & \cite{armanious2020medgan} & - &
	 	{\begin{tabular}[c]{@{}c@{}} \cite{zhang2018harmonic,siddiquee2019learning},\\
	 			\cite{armanious2019unsupervised}\end{tabular}}& -& -& - & \cite{Lee_2019_CVPR,Siddiquee_2019_ICCV} &- & - &- &-\\\hline
	 	dose calculation& -& - & \cite{kaji2019overview}& -& -& -& -& - &-  &- &-\\\hline
	 	{\begin{tabular}[c]{@{}c@{}}surigical training\\ phantoms improvement\end{tabular}}& -& - &\cite{engelhardt2018improving}  & -& -& -& -& -& - &- &- \\\hline
	 	transfer learning& -& - & \cite{gamrian2019transfer}& -& -& -& -& -& - &- &- \\\hline
	 	image registration& -& - &\cite{Arar_2020_CVPR}& -& -& -& -& -& - &- &- \\\hline
	 	domain adaptation& -& - &\cite{Murez_2018_CVPR,cao2018dida} & -& -& - &\cite{liu2018unified}  & -& - &- &-\\\hline
	 {\begin{tabular}[c]{@{}c@{}}person\\re-identification\end{tabular}} & - &-  & {\begin{tabular}[c]{@{}c@{}}\cite{Deng_2018_CVPR,Zhong_2018_CVPR},\\
	 		\cite{Zhong_2018_ECCV,Zheng_2019_CVPR},\\
	 		\cite{Liu_2019_reid}\end{tabular}}& -& -& -& -& -& - &- &- \\\hline
	 	image segmentation & 
	 			{\begin{tabular}[c]{@{}c@{}}\cite{isola2017image,yang2018mri},\\
	 			\cite{GUO2020127,LI2020107343}\end{tabular}}& - & {\begin{tabular}[c]{@{}c@{}}\cite{zhu2017unpaired,lin2020tuigan},\\
 				\cite{li2019asymmetric}\end{tabular}}
 			 & -& -& -& -& -& - &- &- \\\hline
	 	{\begin{tabular}[c]{@{}c@{}}facial geometry\\ reconstruction \end{tabular}} & \cite{Sela_2017_ICCV} & -& -& -& -& -& - &- &- &- &-\\\hline
	 	{\begin{tabular}[c]{@{}c@{}}3D pose\\ estimation\end{tabular}} &- &-  &\cite{Tung_2017_ICCV,Li_2020_CVPR}& -& -& -& -& -& - &- &-\\\hline
	 	{\begin{tabular}[c]{@{}c@{}}neural talking\\ head generation\end{tabular}}  &-  &-  &\cite{Zakharov_2019_ICCV}& -& -& -& -& -& - &-&- \\\hline
		Audio-to-image & \cite{9412890}& -& -& -& - &- & -& - &- &-&-\\\hline
	{\begin{tabular}[c]{@{}c@{}}hand gesture- \\ to-gesture translation \end{tabular}}&\cite{10.1145/3240508.3240704} &- & -& -& -& - &-& - & -&- &-\\\hline		
	\end{tabular}}
\end{table*}

For realistic-looking image synthesis, the related I2I works tend to generate photos of real-world scenes given different forms of input data. A typical task involves translating a semantic segmentation mask \cite{Park_2019_CVPR,Zhu_2020_CVPR,lee2020maskgan,Tang_2020_CVPR,10.1145/3394171.3416270, tang2020edge} into real-world images, that is, semantic synthesis. Person image synthesis, including virtual try-on \cite{10.1145/3123266.3123277,NIPS2017_6644,siarohin2018deformable,Ma_2018_CVPR,Esser_2018_CVPR,Dong_2019_ICCV,Balakrishnan_2018_CVPR, Zhu_2019_CVPR,liu2019liquid,tang2020xinggan,tang2020bipartite} and skeleton/keypoint-to-person translation \cite{zhang2020cross,zhou2020full},~\cite{tang2019cycle} learns to translate an image of a person to another image of the same person with a new outfit as well as diverse poses by manipulating the target clothes or poses. In addition, sketch-to-image translation \cite{isola2017image,Wang_2018_ECCV,wang2018high,bansal2018pixelnn,Chen_2020_CVPR,kazemi2018unsupervised,pmlr-v80-almahairi18a,huang2018multimodal,chen2018sketchygan,lee2018diverse},~\cite{8756586} text-to-image translation \cite{ma2018gan,mao2019mode,yu2019multi,lin2019zstgan},~\cite{tao2020df,Li_2020_CVPR}, audio-to-image translation~\cite{9412890} and painting-to-image translation \cite{park2020swapping,tomei2019art2real,lee2018diverse,benaim2018one} aim to translate human-drawn sketches, text, audio and artwork paintings to realistic images of the real world. Before I2I, most methods relied on the retrieval of existing photographs and copying the image patches to the corresponding location in an inefficient and time-consuming manner. 

Using I2I for image manipulation focuses on altering or modifying an image while keeping the unedited factors unchanged. Semantic manipulation tries to edit the high-level semantics of an image, such as the presence and appearance of objects (image composition) \cite{Shocher_2020_CVPR} with or without makeup \cite{chang2018pairedcyclegan}. Attribute manipulation \cite{8718508,Wu_2019_ICCV,Siddiquee_2019_ICCV} varies the binary representations or utilizes the landmark to edit image attributes, such as the gender of the subject \cite{NIPS2017_7178}, the color of hair \cite{cho2019image,choi2020stargan}, the presence of glasses \cite{press2018emerging} and the facial expression~\cite{8803654,8578838,10.1145/3240508.3240612}, and performs image relabeling \cite{Shocher_2020_CVPR} as well as gaze correction and animation in the wild~\cite{10.1145/3394171.3413981}. Moreover, image/video retargeting \cite{Bansal_2018_ECCV} enables the transfer of sequential content from one domain to another while preserving the style of the target domain.
Much of the I2I research focuses on filling in missing pixels, i.e, image inpainting \cite{Pathak_2016_CVPR,zhu2017toward,Song_2018_ECCV,Liu_2019_ICCV,Zhao_2020_CVPR} and image outpainting \cite{Zhang_2020_CVPR}, but they treat different occluded images. Taking the image of a human face as an example, the image inpainting task produces visually realistic and semantically correct results from the input with a masked nose, mouth and eyes, while the image outpainting task translates a highly occluded face image that only has a nose, mouth and eyes.

I2I has made great contributions to artistic creation. In the past, redrawing an image in a particular form of art requires a well-trained artist and much time. In contrast, many I2I studies can automatically turn photo-realistic images into synthetic artworks without human intervention. Using the I2I methods for artistic creation can directly translate real-world photographic works into illustrations in children’s books \cite{hicsonmez2020ganilla}, cartoon images \cite{DBLP:conf/iclr/TaigmanPW17,Yi_2017_ICCV,kim2019u,gokaslan2018improving,zhao2020unpaired,ma2018gan,shi2019warpgan,wang2020learning,chen2018cartoongan,zheng2019unpaired,lee2018diverse,chang2020domainspecific,lin2020tuigan}, comics \cite{pkesko2019comixify,su2021mangagan} or a multichirography of Chinese characters \cite{gao2020gan}. Additionally, the style transfer task achieves remarkable success through I2I methods. It contains two main objectives: artistic style transfer \cite{zhu2017unpaired,kim2017learning,yi2017dualgan,liu2017unsupervised,lee2018diverse,huang2018multimodal,cho2019image,lee2020drit++,jiang2020tsit}, which involves translating the input image to the desired artistic style, such as that of Monet or van Gogh; and photo-realistic style transfer \cite{Yi_2017_ICCV,park2020swapping,zhu2017unpaired,liu2017unsupervised,jiang2020tsit,park2020contrastive,shen2019towards,chang2020domainspecific}, which must clearly maintain the original edge structure when transferring a style.

We can also exploit I2I for image restoration. The goal of image restoration is to restore a degraded image to its original form via the degradation model. Specifically, the image super-resolution task \cite{Yuan_2018_CVPR_Workshops,zhang2019multiple} involves increasing the resolution of an image, which is usually trained with down-scaled versions of the target image as inputs. Image denoising \cite{manakov2019noise,armanious2020medgan,touvron2020powers} aims to remove artiﬁcially added noise from the images. Image deraining \cite{zhang2019image,li2019heavy,zhu2019singe}, image dehazing \cite{engin2018cycle,cho2019dehazegan,chen2019image,dudhane2019ri,cho2020underwater} and image deblurring \cite{madam2018unsupervised,kupyn2018deblurgan,Liu_2018_ECCV_Workshops,kupyn2019deblurgan} aim to remove optical distortions from photos that were taken out of focus or while the camera was moving, or from photos of faraway geographical or astronomical features. 

Image enhancement is a subjective process that involves heuristic procedures designed to process an image to satisfy the human visual system. I2I proves its effectiveness in this field including image colorization and image quality improvement. Image colorization \cite{zhang2017real,suarez2017infrared,he2018deep,zhang2019deep,xu2020stylization,Lee_2020_CVPR} involves imagining the color of each pixel, given only its luminosity. It is trained on images with their color artiﬁcially removed. Image quality improvement \cite{zhu2017unpaired,ignatov2017dslr,chen2018deep,de2018fast} focuses on producing noticeably fewer colored artifacts around hard edges and more accurate colors, as well as reduced noise in smooth shadows. Moreover, \cite{guo2019fusegan} Learns to fuse multi-focus image using I2I methods.

We also notice that two special types of data are used in I2I algorithms for particular tasks: remote sensing imaging for wildlife habitat analysis \cite{zheng2019exploiting} and building extraction~\cite{ding2021adversarial}; medical imaging for disease diagnosis \cite{zhang2018harmonic,siddiquee2019learning,armanious2019unsupervised,armanious2020medgan}, dose calculation \cite{kaji2019overview} and surgical training phantoms improvement \cite{engelhardt2018improving}. 

I2I methods can also contribute to other visual tasks, such as transfer learning for reinforcement learning \cite{gamrian2019transfer}, image registration \cite{Arar_2020_CVPR}, domain adaptation \cite{Murez_2018_CVPR,cao2018dida,liu2018unified}, person re-identification \cite{Deng_2018_CVPR,Zhong_2018_CVPR,Zhong_2018_ECCV,Zheng_2019_CVPR,Liu_2019_reid}, image segmentation \cite{yang2018mri,GUO2020127,LI2020107343}, facial geometry reconstruction \cite{Sela_2017_ICCV}, 3D pose estimation \cite{Tung_2017_ICCV,Li_2020_CVPR}, neural talking head generation~\cite{Zakharov_2019_ICCV} and hand gesture-to-gesture translation~\cite{10.1145/3240508.3240704}.

\section{Summary and Outlook}
\label{sec:conclusion}
In recent years, the image-to-image translation (I2I) task has achieved great success and benefited many computer visual tasks. I2I is attracting increasing attention because of its wide practical application value and scope. We therefore conduct this comprehensive review of the analysis, methodology, and related applications of I2I to clarify the main progress the community has made. In detail, we first briefly introduce the two most representative generative models that are widely used as the backbone of I2I and some well-known evaluation metrics. Then, we elaborate on the methodology of I2I regarding two-domain and multi-domain tasks. In addition, we provide a thorough taxonomy of the I2I applications.

Looking forward, there are still many challenges in I2I, which need further explorations and investigations. The most iconic dilemma is the trade-off between network complexity and result quality with higher resolution. Similarly, the efficiency should also be considered when the I2I framework attempts to generate diverse and high-fidelity outputs. We believe that a more lightweight I2I network would attract more attention for practical application. Moreover, it is an interesting research trend to generalize the aforementioned methods to domains beyond images, such as those of language, text and speech, termed cross-modality translation tasks. Overall, we hope that this article can serve as a basis for the development of better methods for I2I and inspire researchers in more domains in addition to images.

%

%
%
%

\ifCLASSOPTIONcaptionsoff
\newpage
\fi



%
\bibliographystyle{IEEEtran}
\bibliography{egbib}



%

%
%
%




\end{document}